\documentclass[12pt,a4paper]{report}

\usepackage{fancyvrb,subfloat,amsthm,amssymb,mathrsfs,setspace,mathtools,amsmath}
\usepackage[T1]{fontenc}
\usepackage{lmodern}
\usepackage{amsmath}
\usepackage{booktabs}
\usepackage{csvsimple}
\usepackage{standalone}
\usepackage{url}
\usepackage{fontenc,graphicx,caption,subcaption,epsfig,float,tabularx}
\usepackage[nottoc]{tocbibind}
\usepackage{algorithmic}
\usepackage[boxruled]{algorithm2e}

\renewcommand{\chaptermark}[1]{\markboth{#1}{}}

\theoremstyle{plain}

\theoremstyle{definition}

\theoremstyle{remark}

\usepackage[margin=0.75in]{geometry}

\begin{document}



\begin{titlepage}
\enlargethispage{3cm}

\begin{center}

\vspace*{0.5cm}

\textbf{\Large SUMMARIZATION AND VISUALIZATION OF LARGE VOLUMES OF BROADCAST VIDEO DATA}\\[2cm]


 A report submitted in partial fulfillment of \\
the requirements for the degree of \\[1cm]

{\bf\Large\ Bachelor of Technology }\\[2cm]


{\large \emph{by}}\\[5pt]
{\large\bf {Kumar Abhishek}}\\[1pt]
{\large {and}}\\[1pt]
{\large\bf {Yogi Ashok Sunil}}\\[2cm]

{\large {Under the guidance of}}\\[1pt]
{\large\bf {Dr. Prithwijit Guha}}\\[2cm]
\includegraphics[height=2.5cm]{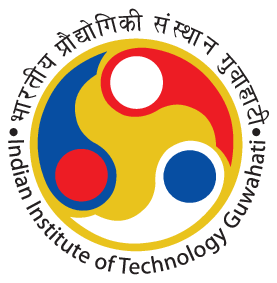}

\vspace*{0.5cm}

{\bf\large \mbox{ DEPARTMENT OF ELECTRONICS}} \\[5pt]
{\bf\large \mbox{\& ELECTRICAL ENGINEERING}} \\[5pt]
{\bf\large \mbox{INDIAN INSTITUTE OF TECHNOLOGY GUWAHATI}}\\[5pt]
{\it\large April 2015}
\end{center}

\end{titlepage}

\clearpage

\pagenumbering{roman} \setcounter{page}{2}
\begin{center}
{\Large{\bf{CERTIFICATE}}}
\end{center}

\noindent
This is to certify that the work contained in this project report
entitled ``{\bf Summarization and Visualization of Large Volumes of Broadcast Video Data}" submitted
by {\bf Kumar Abhishek} and {\bf Yogi Ashok Sunil} to Department of Electronics and Electrical Engineering, Indian Institute of Technology Guwahati under my supervision and that it has not been submitted elsewhere for a degree.

\vspace{4cm}

\noindent Guwahati - 781 039 \hfill Dr. Prithwijit Guha\\
\noindent $ 22^{\text{nd}}$ April, 2015  \hfill (Assistant Professor, EEE Dept.)\\

\clearpage

\addcontentsline{toc}{chapter}{Acknowledgement}%
\begin{center}
{\Large{\bf{ACKNOWLEDGEMENT}}}
\end{center}
We would like to express our deepest appreciation to our thesis supervisor, \textbf{Dr. Prithwijit Guha} who constantly motivated us to work in this direction and valued our ideas. We would like to thank him for the freedom he gave us to carry out research in the field of our interest. We are indebted to \textbf{Mr. Raghvendra Kannao} for the constant support and technical help. We would also like to thank \textbf{Mr. Rajul Gupta} and \textbf{Mr. Rajesh Ratnakaram} for their valuable suggestions. Finally, we would like to thank our parents and friends for their immense love and support during our entire student life.
\clearpage 

\addcontentsline{toc}{chapter}{Abstract}%
\begin{center}
{\Large{\bf{ABSTRACT}}}
\end{center}

\emph{Over the past few years there has been an astounding growth in the number of news channels as well as the amount of broadcast news video data. As a result, it is imperative that automated methods need to be developed in order to effectively summarize and store this voluminous data. Format detection of news videos plays an important role in news video analysis. Our problem involves building a robust and versatile news format detector, which identifies the different band elements in a news frame. Probabilistic progressive Hough transform has been used for the detection of band edges. The detected bands are classified as natural images, computer generated graphics (non-text) and text bands. A contrast based text detector has been used to identify the text regions from news frames. Two classifers have been trained and evaluated for the labeling of the detected bands as natural or artificial - Support Vector Machine (SVM) Classifer with RBF kernel, and Extreme Learning Machine (ELM) classifier. The classifiers have been trained on a dataset of $6000$ images ($3000$ images of each class). The ELM classifier reports a balanced accuracy of $77.38$\%, while the SVM classifier outperforms it with a balanced accuracy of $96.5$\% using $10$-fold cross-validation. The detected bands which have been fragmented due to the presence of gradients in the image have been merged using a three-tier hierarchical reasoning model. The bands were detected with a Jaccard Index of  $0.8138$, when compared to manually marked ground truth data. We have also presented an extensive literature review of previous work done towards news videos format detection, element band classification, and associative reasoning.}

\clearpage

\tableofcontents
\clearpage

\listoffigures
\clearpage

\newpage

\pagenumbering{arabic}
\setcounter{page}{1}


\chapter{Introduction}
\vspace{-0.1in}
Classification of broadcast news videos is the first step towards news content analytics and understanding. The exponentially increasing amount of digital news broadcast videos makes it practically impossible to manually classify the news videos. This demands for an automatic and efficient method of classifying news videos. Various techniques of classifying and analyzing the news videos had been proposed in the past \cite{757471}\cite{Luse}. Most of these methods use prior knowledge about the news format and hence fails to classify the videos whose format is unknown. \\
Format detection is the detection of relevant element bands from a video frame viz. anchor shot, field shot, news ticker, channel logo, news headline etc. and create a format profile using these bands. Arguably, format detection is a precursor for any video content analysis and hence the result of format detection directly affects the efficiency of further video analysis and classification. 
 
  \begin{figure*}[!htb]
    \begin{center}
    \begin{subfigure}[b]{.5\linewidth}

        \includegraphics[scale=0.33]{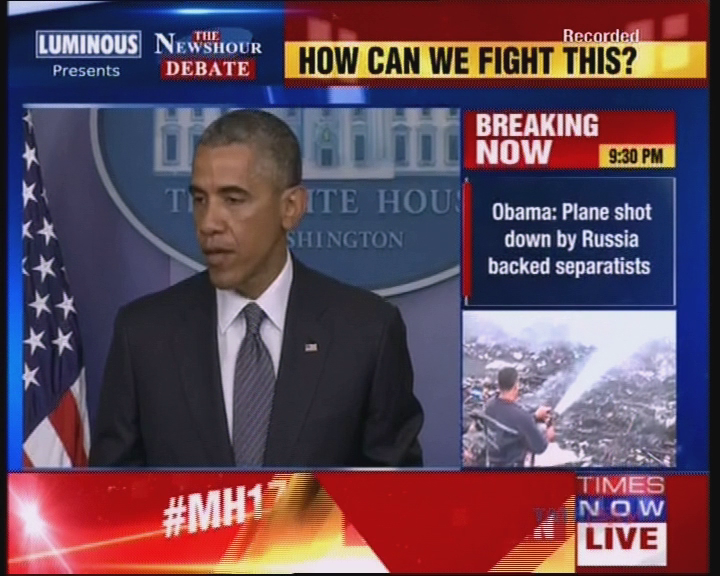}
        \caption{}\label{(a)}
     \end{subfigure}%
     \begin{subfigure}[b]{.5\linewidth}
         \includegraphics[scale=0.4377]{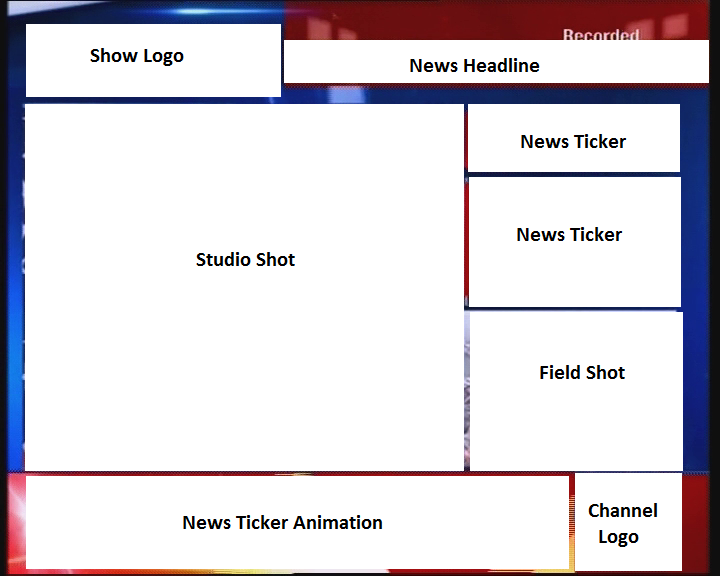}                         

         \caption{}\label{(b)}
      \end{subfigure} \\%

       \caption{Example of news format detection for ``The Newshour Debate" show (a) Original news video frame (b) Expected output format of the news video frame.} \label{fig:eps}
    \end{center}
\end{figure*}

\section{Motivation}
\begin{figure}[H]
\centerline{\includegraphics[scale=0.85]{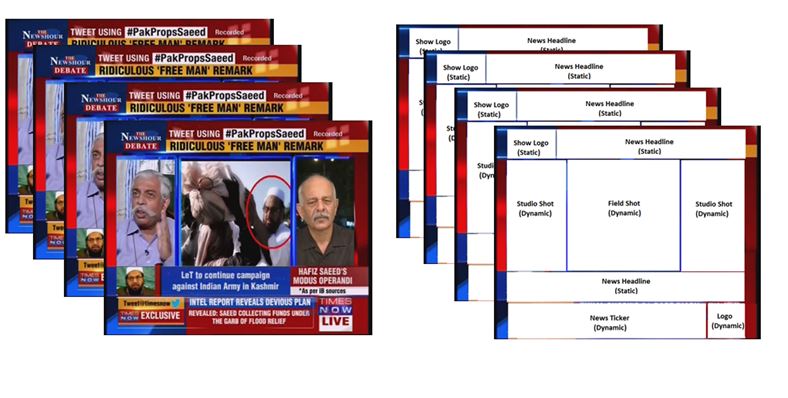}}
\caption{Video frame shots illustrating Dynamic and Static bands.}
\label{fig:dyna}
\end{figure}

Multimedia streaming requires a large bandwidth, especially if the content being transmitted is of very high quality. One page of text can require 500 to 1000 bytes \cite{try} while a single full screen image may require 3,00,000 to 10,00,000 bytes \cite{try2}. These figures indicate that text can be transferred very effectively compared to the images. Hence, transmitting machine-encoded news ticker text instead of the text image can remarkably reduce the bandwidth requirement. Most of the news video frame comprises of individual element bands. Dynamic bands are the ones which gets changed in each frame while static bands are the ones which are constant for a specific time interval Fig.~\ref{fig:dyna}. Transmission of only the dynamic bands in the news video can help in further lowering the bandwidth requirement.

Storage requirement is another constraint which could be reduced by storing only the relevant bands from the video. Detection of irrelevant bands like advertisement ticker and images could be done after detecting the profile format and these bands can then be eliminated while storing the news videos for future reference. Finally, format detection can also play an important role in automatic censoring of the videos.

\section{Proposed Approach}
  In this project we aim to build a robust news format detector which can accurately split a news video frame into different element bands and create a format profile using these bands.\\
   
\begin{figure}[H]
\centerline{\includegraphics[width=0.85\textwidth]{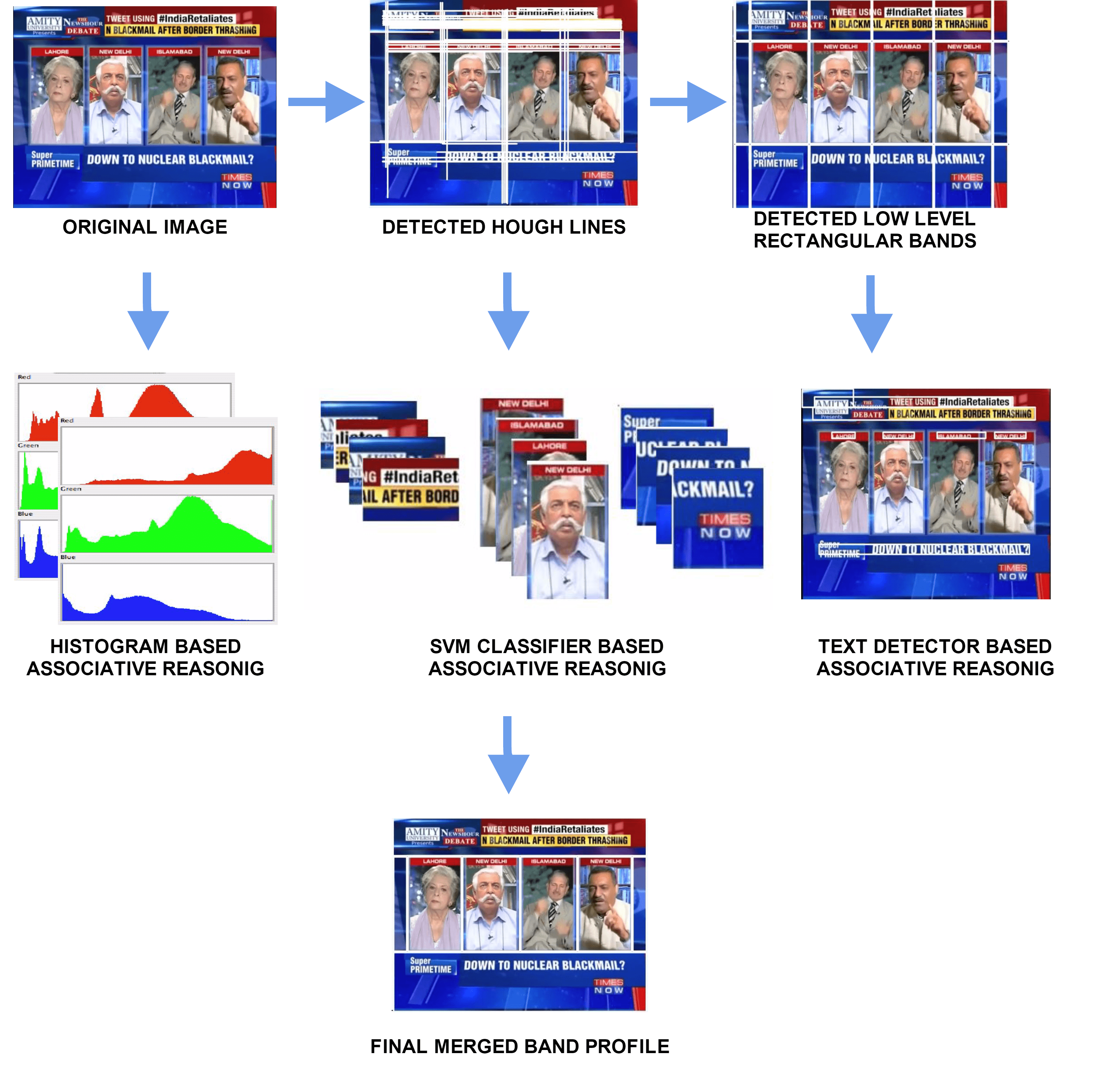}}
\caption{Illustrating the proposed approach for news video format detection.}
\label{fig:approach}
\end{figure}

  Fig.~\ref{fig:approach} summarizes the format detection scheme. Firstly, Hough Transform has been used to detect lines from the video frame. The Hough lines are then extended to the frame boundaries and all the intersection point of the lines are used to generate low-level rectangles. A contrast based text detector \cite{ragh} is used to detect the text bands and a SVM classifier is used to classify the natural and artificial bands. The similar low-level rectangular bands are further merged using the classifier's output to generate the final high-level bands. Finally, these bands are combined together using three-tier associative reasoning to form a format profile.\\

\section{Outline}
The next chapter throws light on the extensive literature review on the methods of news format detection and band classification. In Chapter $3$, we describe the generation of lines using Hough transform and the formation of rectangular bands. Chapter $4$ describes the text-detector for detecting the text regions in the image. In Chapter $5$, we discuss the features used for designing the band classifier. Chapter $6$ elaborates the efficiency of the two classifiers used. In Chapter $7$, we propose a change detection based approach to detect the band type. Chapter $8$ discusses the three-tier reasoning model used for merging the low-level bands. In Chapter $9$ we have mentioned the accuracy of various sub-models and the overall approach. Chapter $10$ concludes the report. Chapter $11$ presents the future work. An Appendix at the end briefly describes the data marking tools created for generating the ground truth data and the training data.

\chapter{Literature Review}
This chapter discusses the literature that provides with the work 
that has been done related with this project. Literature review is 
divided into two parts, namely, Element Band Detection and 
Element Band Classification.
\section{Element Band Detection}
Most of the work in the field of news video analytics and summarization is based on prior knowledge of the video format \cite{Merlino}\cite{4270251}. 
In some of the news video analysis method \cite{4389003}\cite{4035750}, video frames as a whole are directly used for analyzing the videos without dividing it into different format bands. 
Zhang et al. \cite{292432} proposed a method for frame detection to be used in automatic parsing of news video. However, the approach required comparing the video frames with pre-defined news format models and classifying it based on the nearest match. As this method is based on the prior knowledge of the format of news show and news channel, it fails to detect the format in cases where the apriori information is unavailable.\\
Hence, to the best of our knowledge, no previous work has been done in detecting the format of unknown news video and without considering any prior knowledge. 

\section{Element Band Classification}
Previous work done to classify images has been primarily with an aim 
to index the images on the web for search engine optimization (SEO) of 
web image search.

Vassilis Athitsos, Michael J. Swain, and Charles Frankel worked on the classification 
of images into photographs and graphics on the World Wide Web \cite{athitsos}. The 
designed classifier was used by WebSeer \cite{webseer}, an image search engine. The 
features incorporated were total number of different colors, the fraction of pixels having 
the prevalent color, farthest neighbor metric, the fraction of pixels with a saturation greater 
than a certain threshold, color histogram metric, farthest neighbor histogram metric, dimension 
ratio metric and smallest dimension metric. Multiple binary decision trees \cite{quinlan} 
were used for classification and the reported accuracy was 91.4\% on a dataset comprising 
of 9833 graphics and 1733 photographs.

Jing Huang et al. introduced a new feature known as ``color correlogram'' for 
the purposes of image indexing and comparison \cite{correlogramPaper}. Ever since, color 
correlograms have been used widely used for the image classification, amongst other purposes. 
The correlogram is a robust feature and is used to utilize the information from
the spatial correlation of colors in an image.

Rainer Lienhart and Alexander Hartmann worked on heirarchical classification of images \cite{lienhart} 
on the web into 2 broad categories, and then each category was divided into two sub-
categories. Images were classified as photographs and graphical images. The photographs 
were then categorized into actual photographs and artificial, but photo-like images. 
The graphical images were categorized into presentation slides/scientific posters and 
comics/cartoons. The features considered for classification were total number of 
different colors after representing each color channel by only 5 bits (32768 colors), 
the prevalent color, fraction of pixels ($f_1$) from the farthest neighbor metric having a 
distance greater than zero, fraction of pixels ($f_2$) with a distance greater than some chosen 
higher threshold, and the ratio $\dfrac{f_2}{f_1}$. These features were trained on 7516 images 
using Adaboost algorithm, and the algorithm achieved an accuracy of 97.69\%.

Yuanhao Chen et al. worked on the classification of images into photographs and graphics 
for the purpose of web image search as well as personal photograph management \cite{yuanhao}. The 
features used were ranked histogram, color moments (mean, standard deviation and skewness) 
in the HSV color space \cite{stricker}, color correlogram \cite{correlogramPaper}, farthest 
neighbor histogram \cite{athitsos} and the relative size of the largest region and/or the 
number of regions of color constancy (or within a certain tolerance threshold) with a relative 
size bigger than a certain threshold. Adaboost learning algorithm was used as the classifier, 
and it was trained over a dataset of 36000 graphics and 35000 natural images. Five-fold cross 
validation \cite{kfold} was used to validate the algorithm and an accuracy of 94.5\% was reported.

Tian-Tsong Ng et al. also worked on the same problem as Leinhart and Hartmann \cite{lienhart}, 
i.e. distinguishing photographic images and graphics, but instead of using low-level image 
features, they looked at the fundamental differences between the two classes arising due to 
their generation process \cite{tian}. They proposed a geometry-based image model to classify 
images into photographic images (PIM) and photorealistic computer graphics (PRCG). The 
overall accuracy reported was 83.5\%.

Fei Wang and Min-Yen Kan worked on NPIC \cite{fei}, an image classification system that uses 
both metadata-based textual features as well as content based image retrieval (CBIR) features 
to classify images into two categories - natural and synthetic. The metadata-based textual 
features include the filename, file extension, comments (in the image metadata header), image 
URL and the page (where the image is located) URL. The visual features include the image 
specifications, such as the image height, width and the X and Y resolutions (number of pixels 
per inch, i.e. `dpi' along the X and Y dimensions) as well as information extracted from the image 
such as the most common color, the fraction of pixels with the most common color, 
fraction of pixels ($f_1$) from the farthest neighbor metric having a 
distance greater than zero, fraction of pixels ($f_2$) with a distance greater than some chosen 
higher threshold, the ratio $\dfrac{f_2}{f_1}$, the L1, L2 and L-$\infty$ distances \cite{rubner} of the image 
histogram from the average histogram of each category. The training dataset consisted of 16900 
images (graphics and photographs). A boosted decision list learner, BoosTexter \cite{schapire} 
was used as the classifier, and 300 rounds of boosting were performed. The final reported 
testing accuracy was 94.5\%.

\section{Qualitative Spatial Reasoning}
\begin{figure}[H]
\centerline{\includegraphics[scale=0.2]{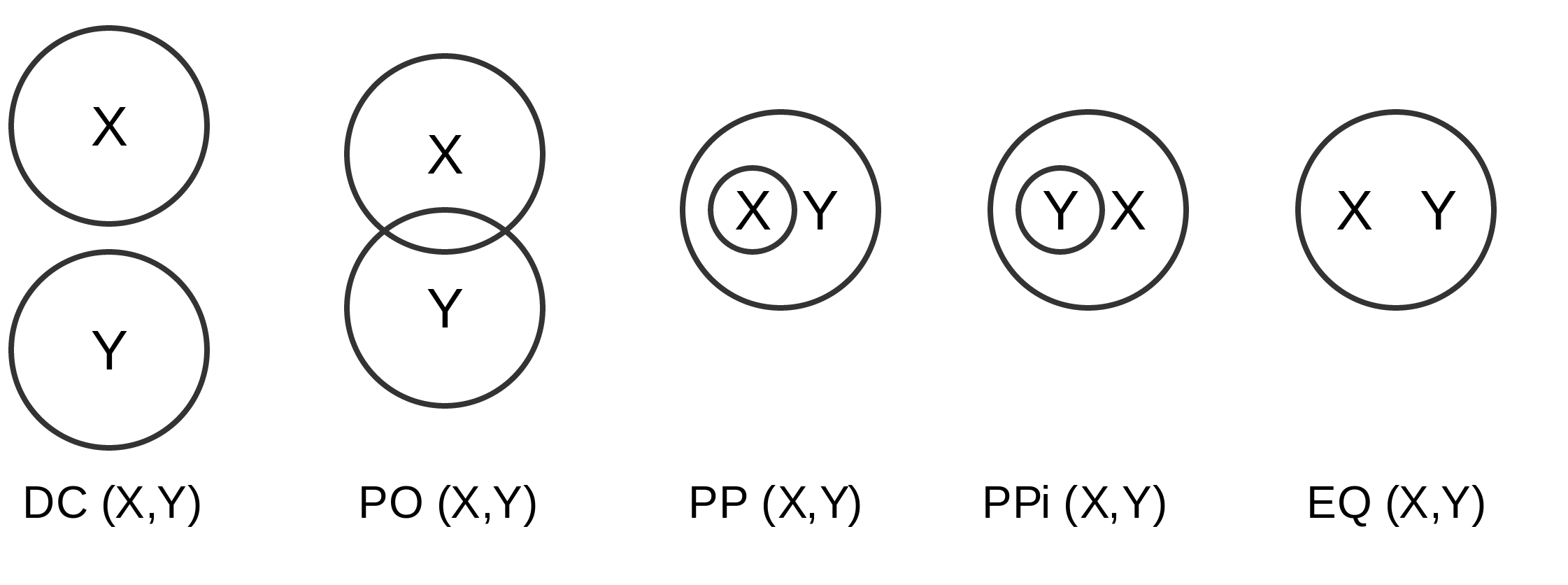}}
\caption{Pictorial representation of the RCC-5 relations.}
\label{rcc5fig}
\end{figure}

Qualitative spatial reasoning \cite{reasoning} is used to represent the relative arrangements of multiple interacting objects. For describing the basic spatial relationship of overlapping objects, the RCC - 5 \cite{rcc} relation proves effective. The RCC - 5 relations shows the extent of overlap between two spatial figures. These five relations, Fig. \ref{rcc5fig} are:
  
\begin{enumerate}
    \item Disconnected (DC):  If $ X \cap Y = \varnothing $
    \item Partial Overlap (PO):  If $ \exists a, b, c: a\in X, a\notin Y, b\in X, b\in Y, c\notin X, c\in Y$
    \item Proper Part (PP):  If $ X \subset Y$
    \item Proper Part Inverse (PPi):  If $ X \supset Y = \varnothing $
    \item Equal (EQ):  If $ X = Y $
\end{enumerate}

Since the RCC - 5 reasoning model is used when the spatial regions are merged together, which is unlikely in our case, a three tier spatial reasoning model is proposed for further reasoning of the detected bands. 

\chapter{Hough Lines and Low-Level Bands Generation}
Most of the bands on the news video frame are made up of basic shapes like square, rectangle or in some cases, trapezium. Hence, to precisely detect these bands, detection of straight lines from the frame is one of the most important steps.\\
Hough transform \cite{Duda} is a method used to isolate defined shapes from an image. The method works by considering the duality between points on the defined curve and its parameters.
Hough transform implements a voting scheme for all potential curves in the image. Each detected edge point votes for the parameter pairs of all the curves to which it belongs. Finally, the curves that have high voting score are the ones which exists in the image.
The Progressive Probabilistic Hough Transform \cite{786993} is the variation of the standard Hough transform which uses the difference in the fraction of votes to precisely detect lines. As a result, progressive probabilistic Hough transform is computationally less expensive compared to the standard Hough transform.\\
The progressive probabilistic Hough transform algorithm suggested by Matas et al. \cite{786993} for detecting lines can be summarized as:

\begin{enumerate}
    \item Check the input image; if it is empty then finish.
    \item Update the accumulator with a single pixel randomly selected from the input image.
    \item Remove the selected pixel from input image.
    \item Check if the highest peak in the accumulator that was modified by the new pixel is higher than threshold \texttt{thr(N)}. If not, goto 1.
    \item Look along a corridor specified by the peak in the accumulator, and find the longest segment that either is continuous or exhibits a gap not exceeding a given threshold.
    \item Remove the pixels in the segment from input image
    \item ``Unvote'' from the accumulator all the pixels from the line that have previously voted.
    \item If the line segment is longer than the minimum length add it into the output list.
    \item Goto 1.
\end{enumerate}

\begin{figure}[!htb]
\minipage{0.32\textwidth}
  \includegraphics[width=\linewidth]{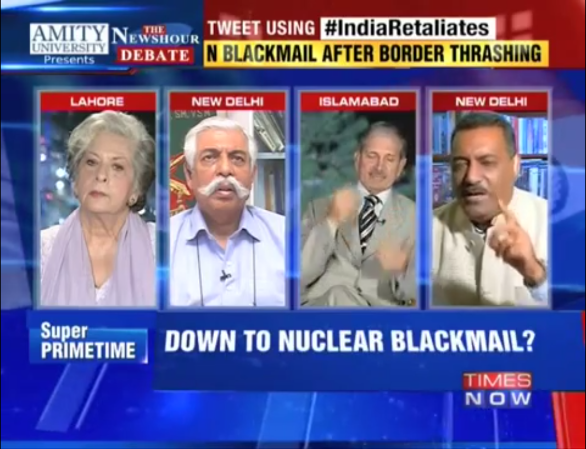}
  \subcaption{Original Image}\label{(a) awesome_image1}
\endminipage\hfill
\minipage{0.32\textwidth}
  \includegraphics[width=\linewidth]{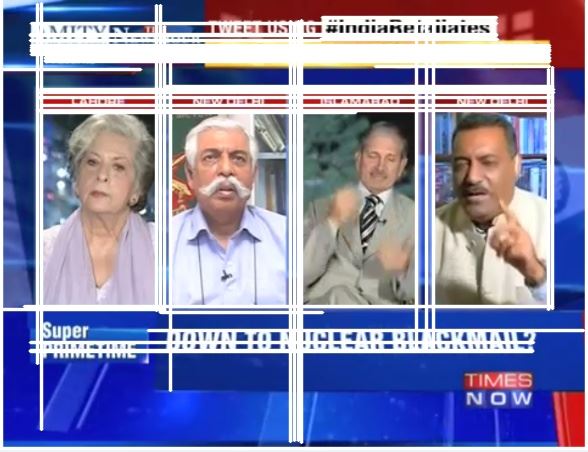}
  \subcaption{Detected Hough Lines}\label{(b) awesome_image2}
\endminipage\hfill
\minipage{0.32\textwidth}%
  \includegraphics[width=\linewidth]{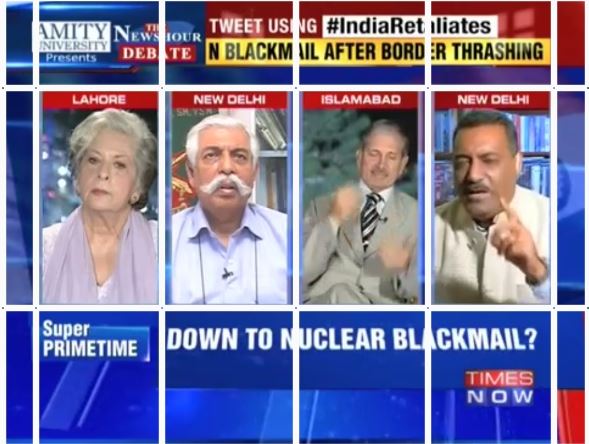}
  \subcaption{Generated Rectangular Bands}\label{(c) awesome_image3}
\endminipage
 \caption{Output of Progressive Probabilistic Hough Transform }
  \label{fig: hoguh}
\end{figure}

The lines detected using PPHT are then extended to image boundaries. This is done to account for the lines which could not be detected due to poor gradient.

The intersection points obtained from the extended Hough lines are then used to form the rectangular bands. These low level rectangular bands divides the entire image into different segmented profiles. In the post-processing step, the smaller profiles from the same class gets merged together to form the final image profile. The final image profile consists of non-overlapping rectangular bands, which defines the profile format which is created by merging the bands using associative reasoning.\\

\chapter{Text Detector}
\label{text_detector}
Raghvendra et. al. \cite{ragh} proposed a two-stage contrast enhancement based pre-processing scheme for the performance improvement of gradient based text detectors, followed by the text band detection using first and second order derivatives of the gradient projection profiles.

 The threshold for gradient magnitude image to generate edge map is a very important precursor for the basic text detection approaches \cite{ragh4}\cite{ragh5}\cite{ragh6}. As stated by Raghvendra \cite{ragh}, it is practically impossible to define a threshold which precisely separates all the text gradient values from those of non text. Due to this, generally a lower threshold is chosen to eliminate definitely non text regions followed by refinement of edge map using text specific features to obtain the text edges from edge map.\\
 Scharr operator \cite{ragh7} is used to compute the normalized gradient magnitude image $ \textit{I$_{m}$} $ due to its isotropic nature. The suppression of small non-zero gradient magnitude is done using the linear contrast enhancement given by:
  \begin{equation}
    I_{mc}(x,y) = \begin{cases}
            \dfrac{1}{\lambda}[\beta]; & \beta>0\\
            0; & Otherwise
            \end{cases}
  \end{equation}

  where $\beta$ = $\alpha($ \textit{I$_{mm}$}(x,y) - 0.5) + 0.5,  $\alpha$ decides the extent of enhancement and  $\lambda$ is the normalization constant\\
  The image $ \textit{I$_{mc}$} $ obtained is further processed with the histogram equilization to obtain $\Omega_{ce}$, which is the contrast enhanced edge map.
  Horizontal Projection Profile $\textit{P$_{hp}$}$ is computed using the contrast enhanced edge map by $\textit{P$_{hp}$(y)}$ = $ \sum_{x=-1}^{width} \Omega(x,y)$ . The horizontal line $y = y*$ is considered to be the member of horizontal text band $ \textit{H$_{band}$}$, if $\textit{P$_{hp}$(y*)}$ exceeds a threshold.\\
  Vertical Projection Profile $\textit{P$_{vp}$}$ is computed within each horizontal band. For the horizontal test band $ \textit{H$_{band}$(y1,y2)}$, $\textit{P$_{vp}$}$ is calculated by $\textit{P$_{vp}$(x)}$ = $ \sum_{y=y1}^{y2} \Omega(x,y)$. Once again, the vertical line $x = x*$ is considered to be the member of vertical text band $ \textit{V$_{band}$}$, if $\textit{P$_{vp}$(y*)}$ exceeds a threshold.\\
  The localization of $ \textit{H$_{band}$(y1,y2)}$ and $ \textit{V$_{band}$(x1,x2)}$ leads to the formation of the bounding text rectangle.\\
  

\begin{figure}[!htb]
\minipage{0.495\textwidth}%
  \includegraphics[width=\linewidth]{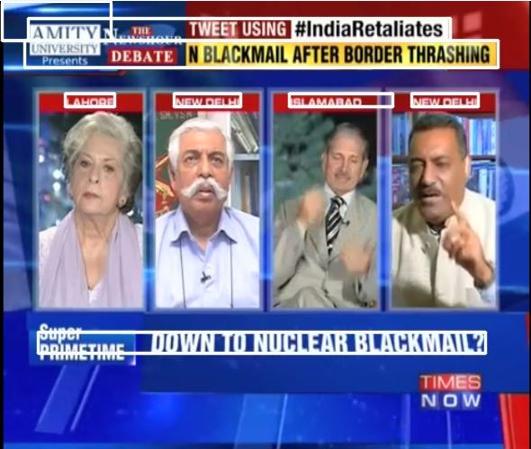}
  
\endminipage\hfill
\minipage{0.495\textwidth}%
  \includegraphics[width=\linewidth]{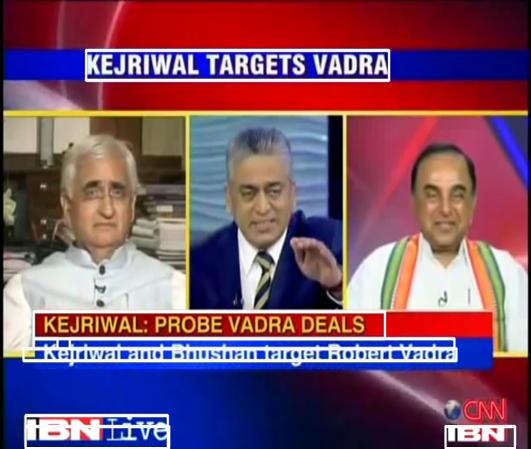}

\endminipage\hfill
\minipage{0.495\textwidth}%
  \includegraphics[width=\linewidth]{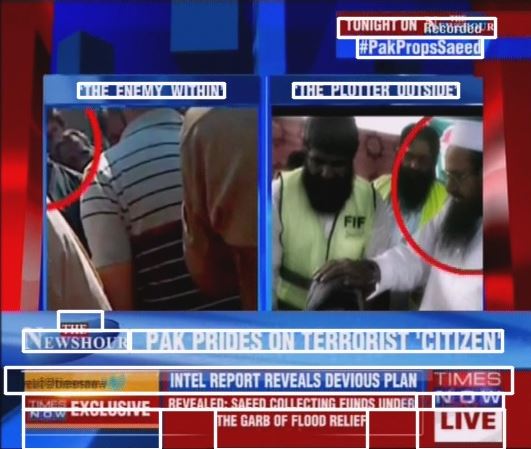}

\endminipage\hfill
\minipage{0.495\textwidth}%
  \includegraphics[width=\linewidth]{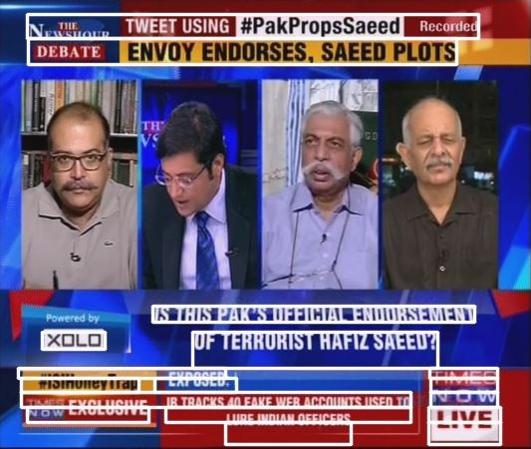}
  
\endminipage\hfill
 \caption{Detected text bands in different news frames  }
  \label{fig: hoguh}
\end{figure}

\chapter{The Feature Space}
After the elements of a frame from a video have been extracted, it is impertive that
they be classified, so as to help in a better understanding of the video format 
discovery. For this purpose, the images have been classified into two categories: 
\textbf{natural images} and \textbf{graphics}. Natural images, as the name 
quite clearly suggests, include images captured by cameras, such as news reporters, 
field shots, etc. Graphics comprise mainly of logo animations, text boxes, ticker 
texts and the news channel logo.
\section{Feature Analysis for Classifier Design}
This section details the features that have been used for the classification
 and also highlights the intuition behind each feature. The chapter has been
 divided into two subsections, corresponding to the color features and the spatial
 features.
\subsection{Color Features}
\label{sec:colorFeatures}
The color features are based on the color information stored in the image pixels. 
Although this may seem quite trivial, the color intensities of the image 
pixels contain a lot of information and the features extracted from them can be
effectively used for the purpose of classification.
\subsubsection{Number of different colors}
\label{subsec:distinctColorScore}
In general, graphics have large areas of constant color, whereas natural photographs don't. As a result, the number of distinct colors in
a graphic image is more than that in a natural photograph. 
However, the number of different colors in an image is not an effective metric to consider as it
 depends on the size of the image. A more accurate metric is the ratio of the number of the different
 colors to the total number of pixels in an image. This score is generally low for graphics as compared to natural images.
 \begin{center}
  $distinctColorScore = \dfrac{\text{Number of different pixels}}{\text{Total number of pixels}}$
 \end{center}

\subsubsection{Prevalent Color Metric}
\label{subsec:prevalentColorScore}
Since graphics tend to have a small number of colors and large regions of color constancy, the number of pixels 
having the most prevalent color is higher in graphics than natural images. Again, we cannot use the exact number of 
pixels for this metric, therefore, we use the ratio of the number of pixels having the most prevalent color 
to the total number of pixels in an image as a classification metric. This score is generally higher for graphics as compared to natural images.
 \begin{center}
  $prevalentColorScore = \dfrac{\text{Number of pixels having the most prevalent color}}{\text{Total number of pixels}}$
 \end{center}

\subsubsection{Saturation Average}
\label{subsec:saturationAverage}
Photographs depict objects of the real world and highly saturated objects are
not very common. Certain colors are much more likely to appear in graphics
than in photographs. For example, animations, maps, charts and logos often
have large regions covered with highly saturated colors. Those colors are
much less frequent in photographs. The saturation average of graphics is 
expected to be greater than that of natural images.\\ \\
Given RGB values, the saturation level of a pixel is defined as the greatest
absolute difference of values between red, green and blue channel intensities. 
The saturation average is then calculted as the average of the saturation levels of 
all the pixels in the image.
 \begin{center}
  $saturationAverage = \dfrac{\sum{max \big(abs(red-blue), abs(blue-green), abs(green-red)\big)}}{\text{Total number of pixels}}$
 \end{center}

\subsubsection{Saturation Metric}
\label{subsec:saturationScore}
This metric is also based on the assumption that highly saturated colors are more 
common in graphics than in photographs. The saturation metric is calculated as 
the ratio of the number of pixels with a saturation level greater then a threshold 
to the total numer of pixels. The saturation metric is expected to be more for the 
graphics than the natural images.
\begin{center}
  $saturationScore = \dfrac{\text{Number of pixels having saturation}\geq\text{THRESHOLD}}{\text{Total number of pixels}}$
 \end{center}

\subsubsection{Color Histogram Metric}
\label{subsec:colorHistMetricScore}  
This metric is based
on the assumption that certain colors occur more frequently 
in graphics than in photographs. In contrast to
the saturation metric, here we do not assume anything
about the nature of those colors. We simply collect
statistics from a large number of graphics and photographs
and construct histograms which show how
often each color occurs in images of each type. The 
score of an image depends om the correlation of its
color histogram to the graphics histogram and the
photographs histogram.\\ \\
A set of 500 images of each class was taken to compute the 
average color histogram of each class. The color histogram metric is 
calculated as the correlation of the image histogram with the average graphics histogram divided 
by the sum of the correlations of the image histogram with the average graphics histogram and 
the average natural image histogram.
\begin{center}
  $colorHistMetric = \dfrac{Graphics\_correlation}{Graphics\_correlation + NaturalImage\_correlation}$
 \end{center}

\subsubsection{Ranked Histogram}
\label{subsec:rankedHist} 
This metric is based on the same assumption as the prevalent color metric: 
graphics tend to have fewer colors than photographs,
and the percentage of the pixels of the prevalent colors for
graphics is higher. However, it is quite possible that an image can contain 
more than one prevalent color. In such cases, the prevalent color metric is 
not very efficient, and the ranked histogram feature prevails.\\ \\
To begin with, each color image (which initially consists of 256 bins for each of 
the three R, G and B channels) is quantized into 32 bins for each channel.
A 32\textsuperscript{3} (=32768) bin histogram is then created, which gives 
the count of each quantized color presented in the image. The histogram is normalized
to unit length in L1 norm. The histogram is sorted
in descending order by the value of the elements. Finally,
the \textbf{m} largest elements are used as the ranked histogram feature.

\subsubsection{HSV Histogram}
\label{subsec:hsv}   
The choice of HSV(Hue Saturation Value) color space instead of RGB can be attributed to the fact that RGB information 
is usually much more noisy than the HSV information. This is because 
unlike RGB, HSV separates \emph{luma}, or the image intensity, from \emph{chroma}, or the color information.\\ \\
Previous work done by Yuanhao Chen et al. \cite{yuanhao} used the moments (mean, standard deviation and skewness) 
of the histogram data from each of the three channels (H, S and V). However, it is not possible 
to encompass all the data using only these three moments. Therefore, it is proposed to use 
the entire 768 bin (256 bins per channel * 3 channels) histogram as a feature for classification 
purpose.

\subsection{Spatial Features}
\label{sec:spatialFeatures}
The color analysis gives some pretty good features that can be used for the classification of images.
However, it does not take in account the spatial correlation of pixels. 
For example, graphics tend to have sharp color gradients, which is not quite the case with natural 
images. Also, because of the way the natural images are acquired, they tend to have some inherent 
noise, which is hardly present in graphics.\\
Different spatial features have been implemented and discussed below.
\subsubsection{Farthest Neighbor Metric}
\label{subsec:farthestNeighbourScore}
This metric is based on the assumption about how color transitions are different
in graphics and natural images. Graphics have comparatively sharp color transitions and large regions
of constant colors as compared to natural images, which have smoother color transitions.\\ \\
For any pixel $p_1$ and its neighbor $p_2$ with their color vectors in the RGB color space 
given as $(r,g,b)$ and $(r',g',b')$ respectively, their color distance \textbf{d} is defined as:
\begin{center}
 $d = \big(abs(r-r') + abs(g-g') + abs(b-b')\big)$
\end{center}

Since the pixel intensity values go from 0 to 255, the value of \textbf{d} can go from 0 to 765.
A threshold between 0 and 765 is specified, and the farthest neighbor score of an image is the 
fraction of pixels having a color distance \textbf{d} greater than or equal to the threshold.
 This score is generally higher for graphics as compared to natural images.
\begin{center}
 $farthestNeighbourScore = \dfrac{\text{Number of pixels for which \textbf{d}}\geq\text{THRESHOLD}}{\text{Total number of pixels}}$
\end{center}

\subsubsection{Farthest Neighbor Histogram Metric}
\label{subsec:farthestNeighbourHistMetricScore}
This metric is based on similar assumptions as the farthest neighbor metric. 
However, this is a different metric as it utilizes the correlation between the
farthest neighbor histogram of the test image with the average farthest 
neighbor histograms of the graphics and the natural images.\\ \\
The farthest neighbor histogram of an image is a one-dimensional histogram with 766 bins. The \emph{i}\textsuperscript{th} bin
 contains the fraction of pixels with transition value equal to \emph{i}.
 The possible of transition values are 0 + 255*3 (for each of the three channels) = 766 total
 transition values.\\ \\
 Using the same dataset of 500 images for each class that was used to generate the 
 average color histogram for the computation of the color histogram metric, we find the 
 average farthest neighbor histogram for each class, and denote them by $FNH_{natural}$ (for 
 natural images) and $FNH_{graphics}$ (for graphics). We define the correlation between two 
 histograms X and Y as 
  \begin{equation}
    C (X, Y) = \displaystyle\sum_{i=0}^{765} X_{i}Y_{i} \nonumber \\
  \end{equation}
where $X_i$ and $Y_i$ are the \emph{i}\textsuperscript{th} bins of X and Y respectively.\\
Let $FNH_{image}$ be the farthest neighbor histogram of a test image. The farthest neighbor 
histogram score of the image is calculated as
\begin{center}
  $farthestNeighbourHistMetricScore = \dfrac{nat}{nat + graph}$
\end{center}

where $nat = C(FNH_{image},FNH_{natural})$ and $graph = C(FNH_{image},FNH_{graphics})$. Similar 
to the farthest neighbor metric, the farthest neighbor histogram metric is expected to be higher 
for the natural images than for the graphics.

\subsubsection{Gray Histogram Smoothness}
\label{subsec:smoothness}
As mentioned in the earlier subsubsections, graphics generally tend to have sharper color 
transitions than natural images. This is also reflected in the gray histogram data.
The graphics histogram shows some narrow and sharp peaks because of the selective use of some 
dominant colors, while this trend is not quite visible on a natural image histogram.\\ \\
The gray histogram smoothness of an image is calculated as: The gray level histogram of the 
image is first calculated. The histogram is then normalized by dividing the value of all 
bins by the number of pixels. The smoothness is the absolute sum of the bin value 
transitions for all bins in the gray histogram.
  \begin{equation}
    smoothness = \displaystyle\sum_{i=1}^{255} \big(grayHist[i]-grayHist[i-1]\big) \nonumber \\
  \end{equation}

\subsubsection{Edge Magnitude Histogram}
\label{subsec:edgeMagHist}
The underlying principle as to why this feature has been incorporated in the classifier 
is that the graphics tend to have strong (prominent) edges, and they are less in number. On the other 
hand, the natural images generally have weaker edges, and they are comparatively more in number.\\ \\
For a given image, its edge magnitude histogram is calculated, and the result is stored 
in 32 bins. This 32-bin histogram is then used as a feature for the classifier.

\chapter{Classification Scheme}
\label{chap:classifier_choice}

\section{Support Vector Machine}
A Support Vector Machine (SVM) with a radial basis function (RBF) kernel has been chosen as 
the classifier, and the SVM parameter estimation was optimised using grid-search with 10-fold 
cross validation \cite{kfold}.

\section{Extreme Learning Machine}
While using feedforward ANNs (Artificial Neural Networks), all the parameters need to be tuned and thus a dependency exists between the parameters (the weights and the biases) of the different layers. Gradient descent based learning methods have been used for learning these parameters, but they are very slow and may give inaccurate results because of convergence to local minima.\\

Proposed by Guang-Bing Huang et al. in 2006, extreme learning machine (ELM) \cite{ELM1}\cite{ELM2}\cite{ELM3} is a learning algorithm for SLFNs (Single Layer Feedforward Networks) having order of magnitudes times faster learning speed and better generalization performance than the classical feedforward network learning algorithms like the Back Propagation Algorithm.\\

\subsection{Theory}
For a generalized SLFN, the hidden layer mapping\footnote{Extreme Learning Machine: Towards Tuning-Free Learning - \url{http://www.ntu.edu.sg/home/egbhuang/pdf/ELM-General.pdf}} can be described as:

\begin{figure}[H]
\centerline{\includegraphics[height=5cm]{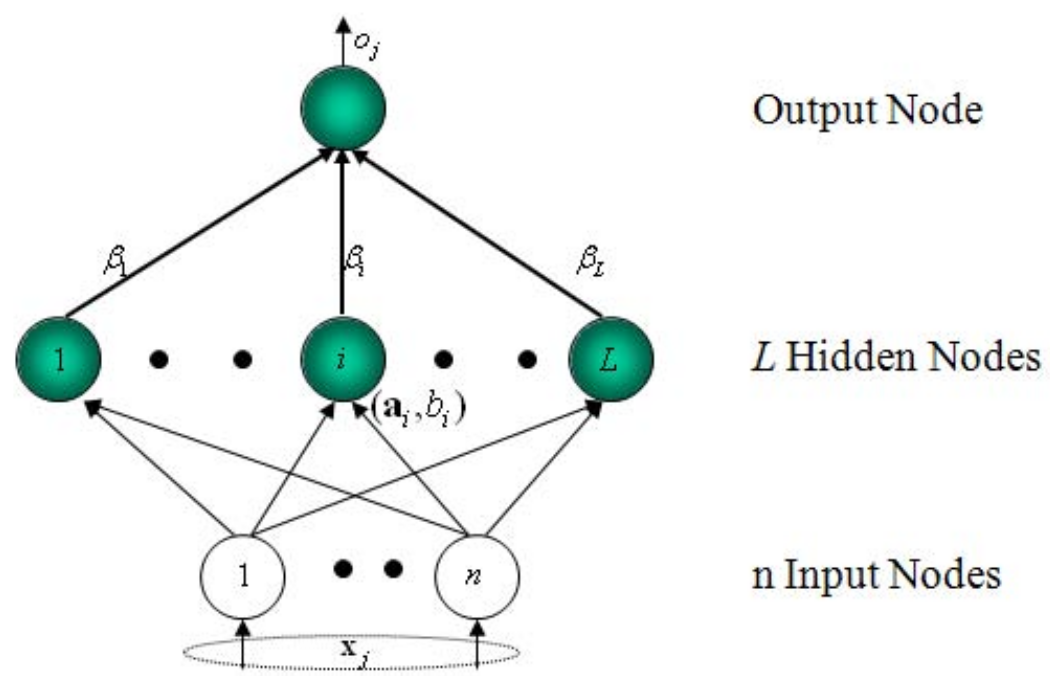}}
\caption{Generalized Single Layer Feedforward Network}
\label{fig:ELM}
\end{figure}

\begin{itemize}
\item Output function of SLFNs:\\
$f_L(\textbf{x}) = \sum_{i = 1}^{L} \beta_{i} G (\textbf{a}_i, b_i, \textbf{x})$
\item The hidden layer output function:\\
$h(\textbf{x}) = [G(\textbf{a}_1, b_1, \textbf{x}), \ldots, G(\textbf{a}_L, b_L, \textbf{x})]$
\item The output functions of the hidden nodes can be but are not limited to:\\
Sigmoid: $G(\textbf{a}_i, b_i, \textbf{x}) = g(\textbf{a}_i \cdot \textbf{x} + b_i)$\\
Radial Basis Function: $G(\textbf{a}_i, b_i, \textbf{x}) = g(b_i \left\Vert \textbf{x} - \textbf{a}_i\right\Vert)$
\end{itemize}
where,\\
$\beta_i$ is the weight vector connecting the $i^{th}$ hidden node and the output nodes, $i = 1, \ldots, L$\\
$\textbf{a}_i$ is the weight vector connecting the input nodes and the $i^{th}$ hidden node, $i = 1, \ldots, L$\\
$b_i$ is the bias vector connecting the $i^{th}$ hidden node and the input nodes, $i = 1, \ldots, L$\\

\begin{algorithm}[H]
\SetAlgoLined
 \textbf{Given}: Training set $\aleph = \{(\textbf{x}_i, \textbf{t}_i) | \textbf{x}_i \in \mathbf{R}^n, \textbf{t}_i \in \mathbf{R}^m, i = 1, \ldots, N\}$, hidden node output function $G(\textbf{a}, b, \textbf{x})$, mumber of hidden nodes $L$\ \\
 \textbf{Assign}: Randomly generated values to the hidden node parameters $(\textbf{a}_i, b_i), i = 1, \ldots, L$.\ \\
 \textbf{Calculate}: The hidden layer output matrix \textbf{H}, the $i^{th}$ column of which is the output of the $i^{th}$ hidden node with respect to the inputs $\textbf{x}_i, \ldots, \textbf{x}_N$.\ \\
 \textbf{Calculate}: The output weight matrix $\beta$ is calculated as $\beta = \textbf{H}^{\dagger}\textbf{T}$, where $\textbf{H}^{\dagger}$ is the \textit{pseudoinverse} of $\textbf{H}$, and $\textbf{T} = [\textbf{t}_1^T, \ldots, \textbf{t}_N^T]^T$, $\textbf{t}_j = \sum_{i = 1}^{L} \beta_{i} G (\textbf{a}_i, b_i, \textbf{x})$.\
 \caption{Extreme Learning Machine Training}
\end{algorithm}

The salient features and advantages of ELMs are as listed below:
\begin{itemize}
\item The learning speed of ELM is extremely fast.
\item The hidden node parameters $\textbf{a}_i$s and $b_i$s are independent of the data and also of each other.
\item ELM does not encounter traditional gradient-based learning issues, such as, local minima, improper learning rate, overfitting etc.
\end{itemize}

\chapter{Band Identification Through Change Detection}
One of the main factors that distinguish artificial bands from the natural bands is that the content of the artificial bands tends to be static over certain range of frames. This is because of the fact that natural bands contain images captures from cameras, such as news reporters, field shots etc., which is updated in each frame while the artificial bands contain computer generated graphics like logo animations, text boxes and ticker texts which remain static over a certain range of frames. This information is used to differentiate between the two bands. Two types of change detection methods were used.

\begin{enumerate}
    \item Pixels based change detection.
    \item Histogram based change detection.
\end{enumerate} 

\section{Pixels Based Change Detection}
For pixels based change detection, a difference image is created for each frame which is the pixel-wise difference between the current frame grey image and the previous frame grey image.

  \begin{equation}
    d^{k}(x,y) = I^{k}(x,y) - I^{k-1}(x,y)
  \end{equation}
  
where, $d_{k}$ is the difference image for the $k$th frame while $I^{k}$ and $I^{k-1}$ are the $k$th and $k-1$th frame respectively. 
Next, the difference image is binarized using a threshold. If the absolute value of difference pixel is greater than the threshold, it is labeled as $1$, otherwise $0$. This binarized image is further divided into $N$x$N$ sub-images and the sub-image is labeled according to the majority of the pixels in that sub-image. 
The natural sub-image is the one with its label as $1$, and the artificial sub-image is one with its label as $0$. 

\section{Histogram Based Change Detection}

For histogram based change detection, first of all,  the image is converted into a monochrome image according to the equation \ref{monochrome}. For the change detection on the $k^{th}$ frame, the $k^{th}$ and the $(k-1)^{th}$ frame are each divided into $N$x$N$ sub-images. Then the histogram is calculated for these sub-images. Corresponding to each sub-image, Bhattacharya distance (Equation \ref{Bhattacharya}) is calculated between the histogram of current frame's sub-image and the histogram of the previous frame's sub-image. If the distance is greater than a given threshold then the sub-image is labeled as natural image, otherwise it is labeled as artificial.\\
\begin{figure}[H]
\centerline{\includegraphics[scale=0.35]{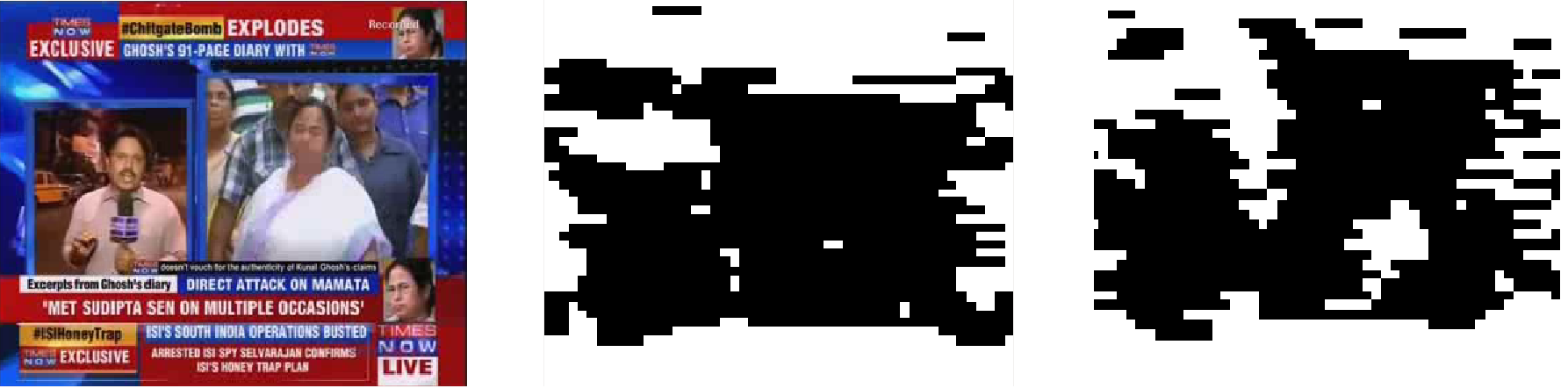}}
       \caption{Output of the change detection (a) Rectanglular bands detected from Hough Lines, (b) Output of the pixels based change detection with N = 50, (c) Output of the histogram based change detection with N = 50.} \label{fig:eps}
\end{figure}
In addition to the classifier output, the change based band identification is used to improve the accuracy and provide better result after associative reasoning.

\chapter{Associative Reasoning}
\vspace{-0.1in}
The low level rectangular bands after the Hough transformation are obtained by using Hough
lines extended till the image boundaries. These bands can be broadly categorized into three categories:
\begin{enumerate}
    \item Synthetic Bands: Synthetic bands are the bands generated artificially using computer graphics. They usually includes logo animations, text boxes, ticker texts and the news channel logo.
    \item Natural Bands: Natural
bands, as the name quite clearly suggests, include bands captured by cameras, such as news reporters, field shots, etc.
    \item Text Bands: Text bands are the bands containing texts. 
\end{enumerate}

Because of the Hough lines extended till the image boundaries, the entire text band gets split into various smaller bands. To compensate for this, the small rectangular bands need to be merged together using associative reasoning. The entire reasoning model can be hierarchically divided into three sub-reasoning models. They are,
  
\begin{enumerate}
    \item Merging of the bands with similar histograms, used for the merging of synthetic bands.
    \item Merging of the bands labeled as natural bands using the SVM classifier. 
    \item Merging of the bands labeled as text bands using the text detector.
\end{enumerate}   
    
For the fast and easy implementation of the associative reasoning, an adjacency matrix is created for each frame. An adjacency matrix $A$, is a $N X N$ square matrix, where $N$ is the total number of bands detected from the frame.

$a_{ij}$ = $ \alpha $ where $ \alpha $ $\in$ $\lbrace$ $0,1,2,3,4$ $\rbrace$. Here, $a_{ij}$ is the element of $i$th row of A and $j$th column of A. 

$a_{ij}$ = $1$ ,if the $j$th band share a boundary with the $i$th band and is above the $i$th band.

$a_{ij}$ = $2$ ,if the $j$th band share a boundary with the $i$th band and is to the right of the $i$th band.
     
$a_{ij}$ = $3$ ,if the $j$th band share a boundary with the $i$th band and is below the $i$th band.

$a_{ij}$ = $4$ ,if the $j$th band share a boundary with the $i$th band and is to the left of the $i$th band.

$a_{ij}$ = $0$ ,if $a_{ij}$ $\notin$ $\lbrace$ $1,2,3,4$ $\rbrace$ , that is, the $i$th and the $j$th band does not share a boundary with each other. 
\section{Merging Rules}

If two text bands $ \textit{H$_{band_i}$}(x_{i},y_{i},h_{i},w_{i})$ with its left corner at $(x_{i},y_{i})$, width $w_{i}$ and height $h_{i}$; and $ \textit{H$_{band_j}$}(x_{j},y_{j},h_{j},w_{j})$ are merged together to form $ \textit{H$_{band_k}$}(x_{k},y_{k},h_{k},w_{k})$. And if,

\begin{enumerate}
  \item $a_{ij}$ = $1$, then
  
  \begin{align*}
    x_{k} &= x_{j}  &   y_{k} &= y_{j} \\
    h_{k} &= h_{i} + h_{j}  &   w_{k} &= w_{j}
  \end{align*}
  \item  $a_{ij}$ = $2$, then
  \begin{align*}
    x_{k} &= x_{i}  &   y_{k} &= y_{i} \\
    h_{k} &= h_{i}  &   w_{k} &= w_{i} + w_{j}
  \end{align*}
  \item  $a_{ij}$ = $3$, then  
  \begin{align*}
    x_{k} &= x_{i}  &   y_{k} &= y_{i} \\
    h_{k} &= h_{i} + h_{j}  &   w_{k} &= w_{i}
  \end{align*}    
  \item  $a_{ij}$ = $4$, then
    \begin{align*}
    x_{k} &= x_{j}  &   y_{k} &= y_{j} \\
    h_{k} &= h_{j}  &   w_{k} &= w_{i} + w_{j}
  \end{align*}
\end{enumerate}

\section{Merging of the bands with similar histograms}
In the first step of associative reasoning, the synthetic bands need to be merged together. This is done by comparing the histograms of two bands. As the synthetic band is uniformally colored or generally does not have much color gradient, its sub-divided bands can be identified using a histogram comparison approach. Adjacent bands having similar histograms are the sub-bands of the same parent synthetic band and hence are merged together.\\

For computing the histogram, firstly, the RGB band image is linearly mapped to have pixel values from $0$ to $32$. Then this RGB image is converted to a monochrome image using the function
  \begin{equation} \label{monochrome}
    I_{mc}(x,y) = 32^{2}*I_{r}(x,y) + 32*I_{g}(x,y) + I_{b}(x,y)
  \end{equation}
  
  where, $I_{mc}$ is the monochrome image and $I_{r}$, $I_{g}$ and $I_{b}$ are the red, green and blue channels of the RGB image respectively. 

Next, normalized histograms are calculated for both the monochrome images. The similarity between both the histograms is measured using the Bhattacharya distance. The Bhattacharya distance between two histograms $H_{1}$ and $H_{2}$ can be calculated as\footnote{OpenCV implementation: \url{http://docs.opencv.org/doc/tutorials/imgproc/histograms/histogram_comparison/histogram_comparison.html}}

  \begin{equation} \label{Bhattacharya}
    d(H_{1},H_{2}) = \sqrt{1-\dfrac{1}{\sqrt{\bar{H_{1}}\bar{H_{2}}N^{2}}} \sum_{I}^{} \sqrt{H_{1}(I).H_{2}(I)}}
  \end{equation}
  
where,

  \begin{equation}
    \bar{H_{k}} = \dfrac{1}{N}\sum_{J}^{}H_{k}(J)
  \end{equation}
  
  Similar histograms have a Bhattacharya distance close to $0$, while two dissimilar histograms have a Bhattacharya distance closer to $1$. Hence, if the two bands share a boundary and have a Bhattacharya distance less than a threshold, then they are merged together according to the merging rules mentioned above.

  \pagebreak
\section{Merging of the natural bands}
A SVM classifier, which is described in Chapter 6 is used to classify the bands into artificial and natural bands. As mentioned before, natural bands are the bands captured by cameras, such as news reporters, field shots, etc and hence sub-divided natural bands needs to be merged to form a complete natural shot frame. Two bands labeled as natural are merged together only if they both share a common boundary but there is no edge present along that common boundary. The bands are merged together according to the merging rules mentioned above.

\section{Merging of the Text bands}
The text detector mentioned in Chapter \ref{text_detector} is used to detect the regions of text in the frame. A low level rectangular band is labeled as text band using the overlap criterion 

  \begin{equation}
    J(R,T) = \dfrac{A(R \cap T)}{A(R)} 
  \end{equation}
  
where, $J(R,T)$ is the text overlap measure of the rectangular band $R$ and text region $T$, $A(R \cap T)$ is the area of intersection of band $R$ and text region $T$, while $A(R)$ is the area of band $R$. 

Two bands with the text overlap measure greater than a given threshold are merged together only if they both share a common boundary and are horizontally adjacent to each other. The bands are merged together according to the merging rules mentioned above.

  \begin{figure*}[!htb]
    \begin{center}
    \begin{subfigure}[b]{.5\linewidth}

        \includegraphics[scale=0.545]{ashok_figs/hough_rects.jpg}
        \caption{}\label{(a)}
     \end{subfigure}%
     \begin{subfigure}[b]{.5\linewidth}
         \includegraphics[scale=0.545]{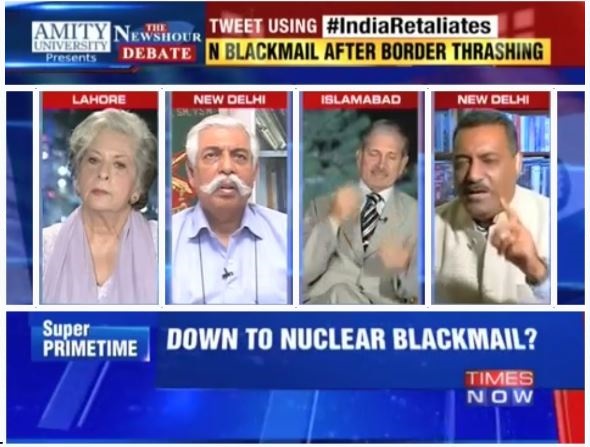}                         

         \caption{}\label{(b)}
      \end{subfigure} \\%
     
       \caption{Output after merging of the bands containig text (a) Rectanglular bands detected from Hough Lines, (b) Output after the application of associative reasoning.}\label{fig:eps}
    \end{center}
\end{figure*}

\chapter{Results}

\section{Element Band Detection}
The format detection approach was tested on 100 images generated from various news channels and different shows to measure its robustness on different types of format profile.\\
Jaccard Index \cite{jaccard} is used as a measure of similarity between the obtained result and the ground truth. For two rectangular bands $R_1$ and $R_2$, Jaccard Index \textbf{\textit{J}}($R_1, R_2$) is defined as
  \begin{equation}
    J(R_1, R_2) = \dfrac{A(R_1 \cap R_2)}{A(R_1 \cup R_2)} \nonumber \\
  \end{equation}

where, $A(R_1 \cap R_2)$ is the area of intersection of $R_1$ and $R_2$ and $A(R_1 \cup R_2)$ is the area of union of $R_1$ and $R_2$. A Jaccard Index of 1 signifies 100\% overlap of the two bands. For the entire image frame $ \textit{I}_{frame} $, the net Jaccard Index is calculated as
 
\begin{equation}
J(\textit{I}_{res},\textit{I}_{gt}) = \dfrac{ \sum_{i=1}^{n_{1}} \max_{1 \leq j \leq n_{2}}( J(R_i, R_j)) }{n_{2}} \nonumber \\
\end{equation}

where, $ \textit{I}_{res} $ is the output of the format detector containing bands $R_i$ for $i=1,2,3 ....n_{1} $ and $ \textit{I}_{gt} $ is the ground truth containing bands $R_j$ for $j=1,2,3 ....n_{2} $\\
The average Jaccard Index of the entire dataset of 100 images was calculated to be $0.8138$.

\section{Element Band Classification}
\label{sec:svmTrain}
The Support Vector Machine (SVM) with a Radial Basis Function (RBF) kernel was 
trained using LibSVM \footnote{LibSVM tool is available at http://www.csie.ntu.edu.tw/~cjlin/libsvm}
tool over a dataset of 6000 images, with equal number of images from both the 
classes. A total of eleven features (color features and spatial features) were 
used for classifier training. The feature vectors were 1320-dimensional each. 
The entire process of feature extraction, SVM training and performance analysis (10-fold cross validation and 
testing) took 530 minutes on an average on an Intel Core i7 Ivy Bridge machine running 64-bit Ubuntu OS with 
8 GB of RAM, when utilizing all the 8 cores in parallel using GNU Parallel \cite{gnuParallel}.\\ \\
The Extreme Learning Machine (ELM) was also trained on the same dataset of 6000 images, each feature vector being 1320-dimensional. The training and performance analysis was comparatively faster than that of SVM.

\section{Performance Measures}
\label{performance}

\subsection{Precision}
Precision is a measure of the accuracy provided that a specific class has 
been predicted. It is defined as:
\begin{center}
 Precision = $\dfrac{\text{TP}}{\text{TP + FP}}$
\end{center}
where \textbf{TP} and \textbf{FP} are the numbers of true positive and 
false positive predictions for the considered class.

\subsection{Recall}
Recall is a measure of the ability of a prediction model to select 
instances of a certain class from a data set. Since we are dealing 
with binary classification here, the recall is also called as sensitivity. 
The recall (or sensitivity) corresponds to the true positive rate, 
and is defined as:
\begin{center}
 Recall = Sensitivity = $\dfrac{\text{TP}}{\text{TP + FN}}$
\end{center}
where \textbf{TP} and \textbf{FN} are the numbers of true positive and 
false negative predictions for the considered class. Thus, (TP + FN) 
corresponds to the total number of test examples of the considered class.

\subsection{F-measure}
The F-measure (also known as $F_1$-score), is a more accurate metric of 
classifier performance than either precision or recall alone. In binary 
classification, the F-measure is the harmonic mean of the precision and 
the recall, as is calculated as:
\begin{center}
 F-measure = 2 $\dfrac{\text{Precision . Recall}}{\text{Precision + Recall}}$
\end{center}

\subsection{Balanced Accuracy}
Evaluating the performance of a classification by averaging the accuracies 
obtained on individual cross-validation folds is not a very good approach \cite{balancedAccuracy} due to two main reasons. 
Firstly, it does not allow for the 
derivation of meaningful confidence intervals. Secondly, it leads to an 
optimistic estimate when a biased classifier is tested on an imbalanced 
dataset. Balanced accuracy, which is defined as the arithmetic mean of the sensitivity 
and the specificity of the classifier, does away with inflated performance estimates on 
imbalanced datasets.
\vspace{0.1in}
\begin{center}
 Balanced Accuracy = $\dfrac{\text{Sensitivity + Specificity}}{2}$ = $\dfrac{\text{0.5*TP}}{\text{TP + FN}} + \dfrac{\text{0.5*TN}}{\text{TN + FP}}$
\end{center}
where \textbf{TP}, \textbf{TN}, \textbf{FP} and \textbf{FN} are the numbers 
of true positive, true negative, false positive and false negative predictions.

\section{Performance Analysis}
\subsection{SVM Performance Analysis}
A graphical representation of the classifier performance for each of the 11 
features has been attached in the next page. A more detailed analysis follows.

\begin{figure}[H]
\centerline{\includegraphics[height=7cm]{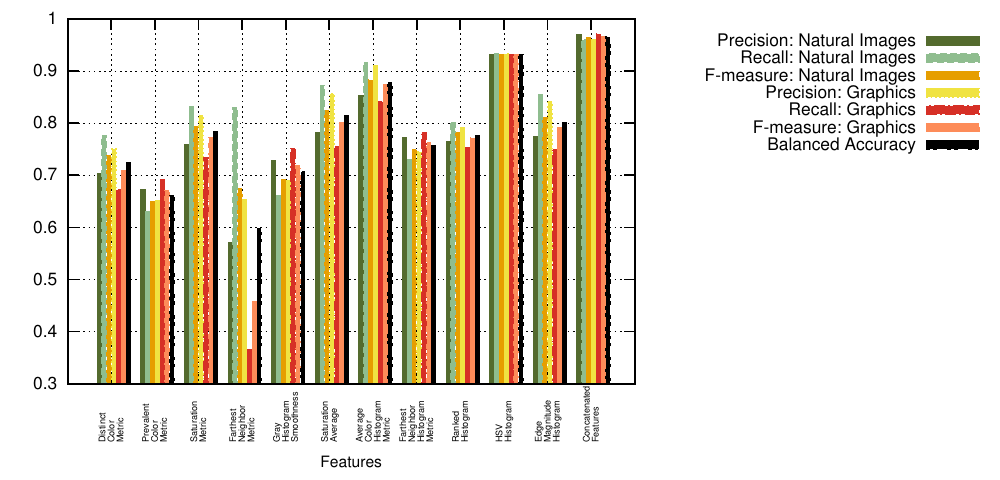}}
\caption{Performance analysis of the designed SVM classifier on individual 
features and their combination. The comparison of precisions, recalls and f-measures for both
natural images and graphics categories are shown. Note that the HSV 
Histogram dominates as a feature.}
\label{fig:AllAnalysis}
\end{figure}

10-fold cross-validation and testing was performed on the dataset to analyse 
the performance of individual features as well as the concatenated features. 
The results have been plotted for all of them and are as shown below:

\begin{figure}[H]
\centerline{\includegraphics[height=7cm]{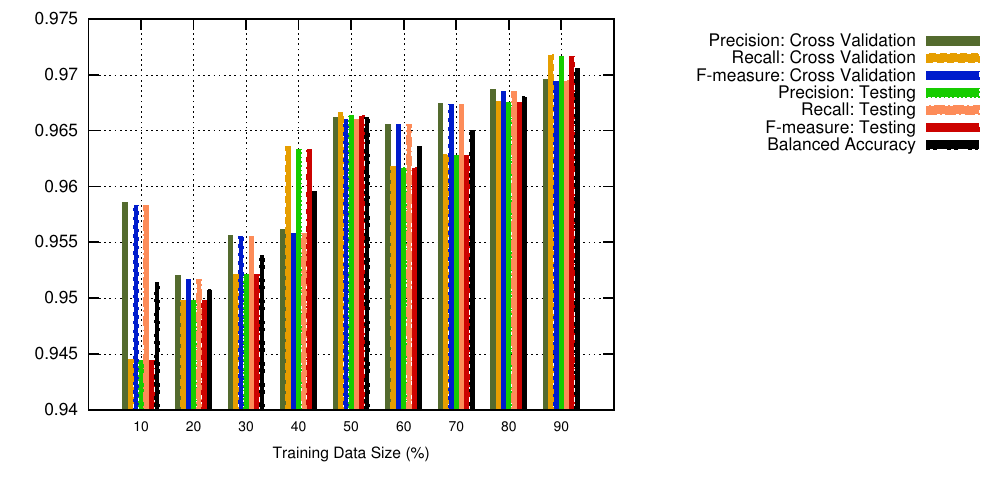}}
\caption{Performance analysis of concatenated features with varying data size}
\label{fig:CatAnalysis}
\end{figure}

\begin{figure}[H]
\centerline{\includegraphics[height=7cm]{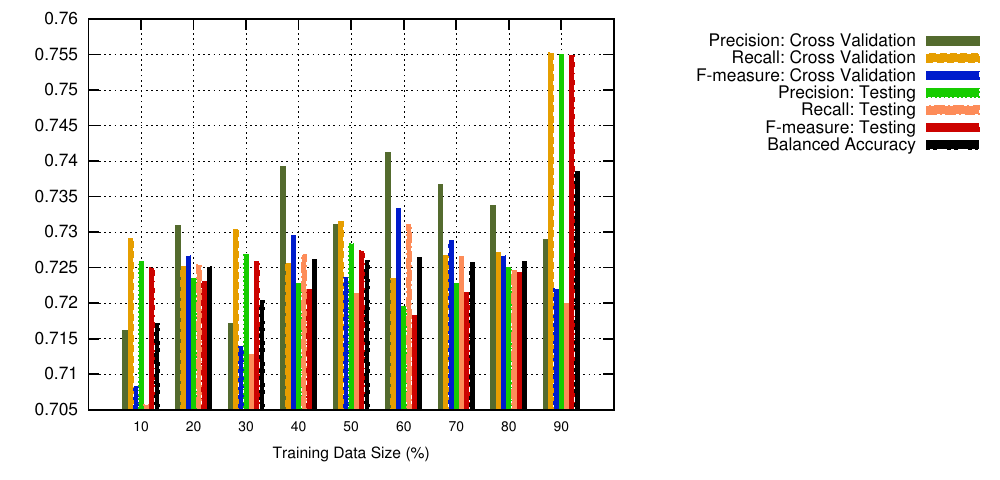}}
\caption{Performance analysis of Distinct Color Metric with varying data size}
\label{fig:CatAnalysis}
\end{figure}

\begin{figure}[H]
\centerline{\includegraphics[height=7cm]{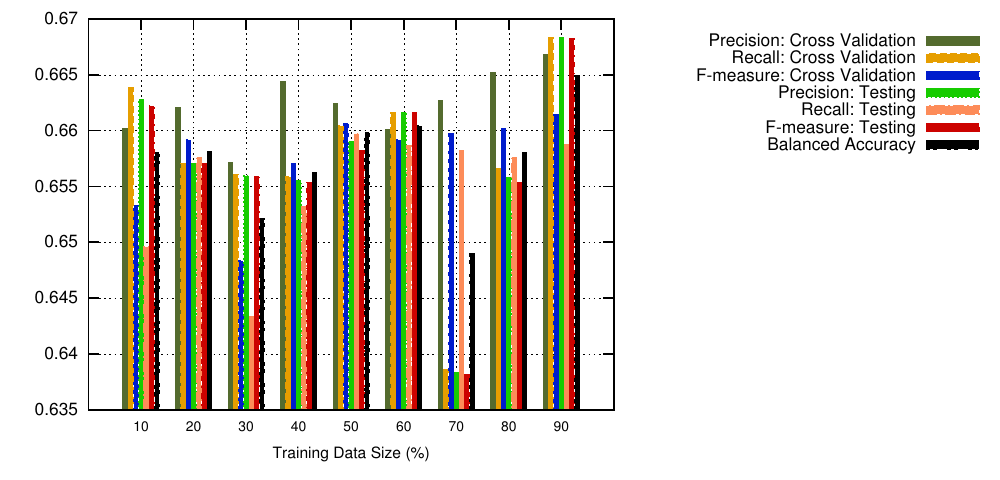}}
\caption{Performance analysis of Prevalent Color Metric with varying data size}
\label{fig:CatAnalysis}
\end{figure}

\begin{figure}[H]
\centerline{\includegraphics[height=7cm]{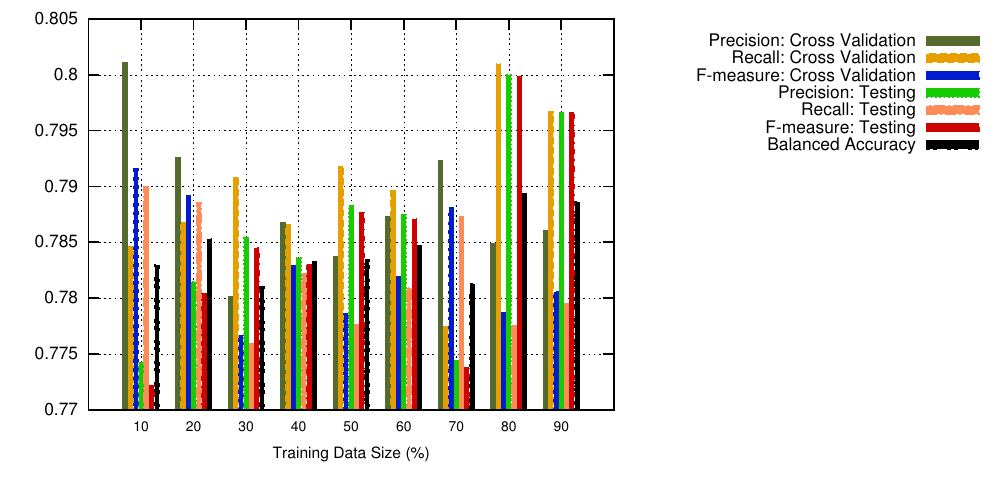}}
\caption{Performance analysis of Saturation Metric with varying data size}
\label{fig:CatAnalysis}
\end{figure}

\begin{figure}[H]
\centerline{\includegraphics[height=7cm]{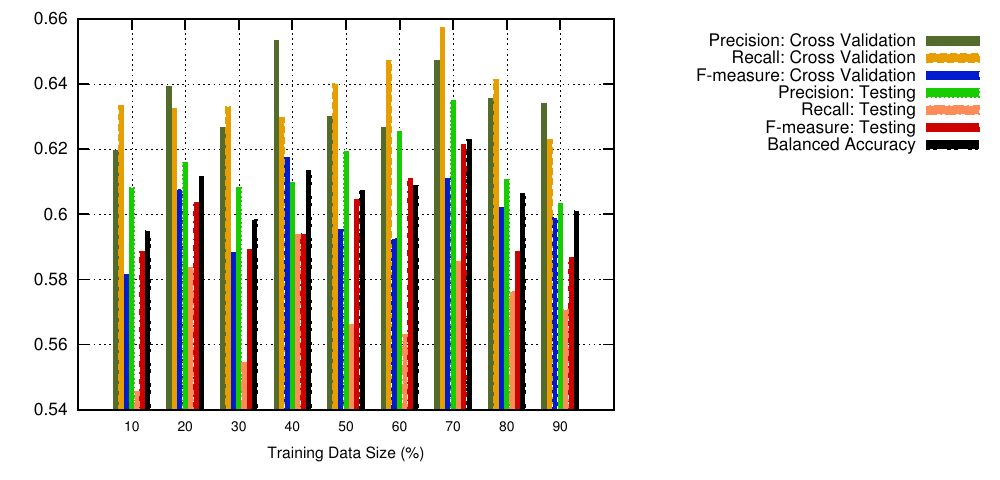}}
\caption{Performance analysis of Farthest Neighbor Metric with varying data size}
\label{fig:CatAnalysis}
\end{figure}

\begin{figure}[H]
\centerline{\includegraphics[height=7cm]{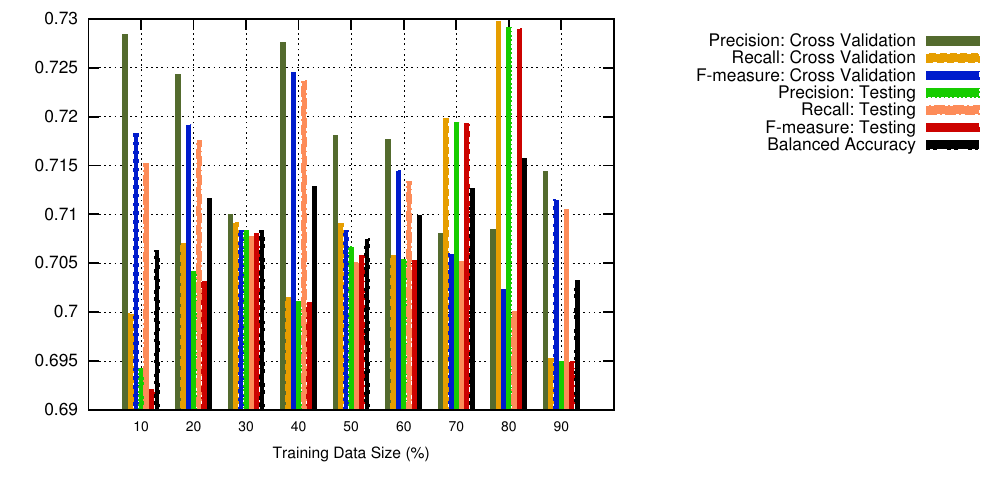}}
\caption{Performance analysis of Gray Histogram Smoothness with varying data size}
\label{fig:CatAnalysis}
\end{figure}

\begin{figure}[H]
\centerline{\includegraphics[height=7cm]{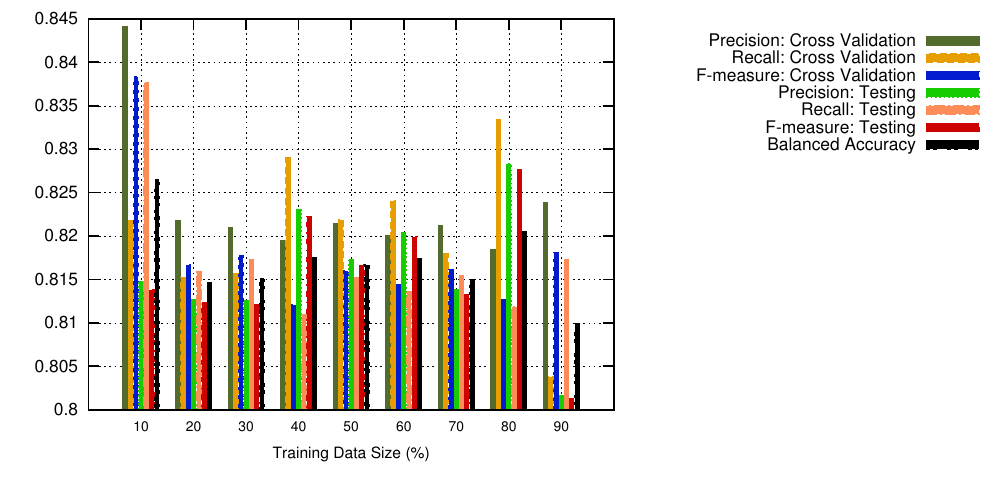}}
\caption{Performance analysis of Saturation Average with varying data size}
\label{fig:CatAnalysis}
\end{figure}

\begin{figure}[H]
\centerline{\includegraphics[height=7cm]{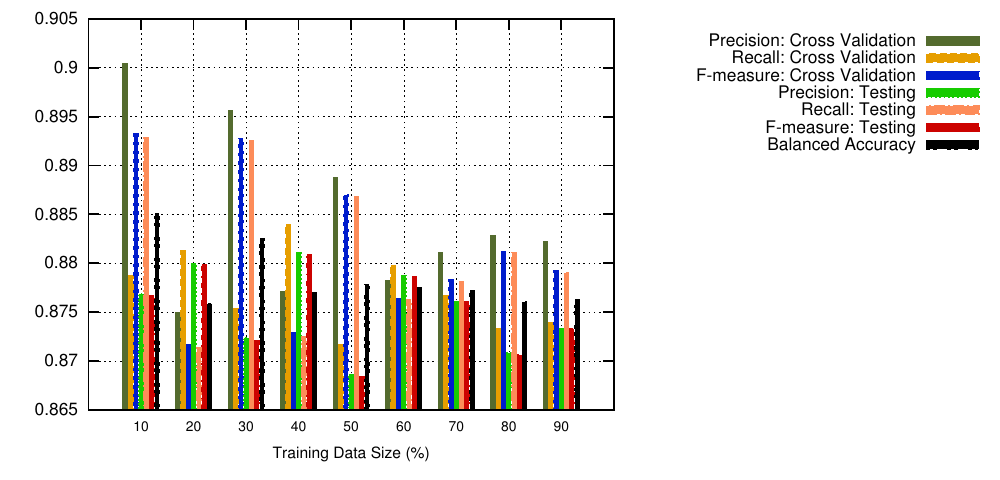}}
\caption{Performance analysis of Average Color Histogram Metric with varying data size}
\label{fig:CatAnalysis}
\end{figure}

\begin{figure}[H]
\centerline{\includegraphics[height=7cm]{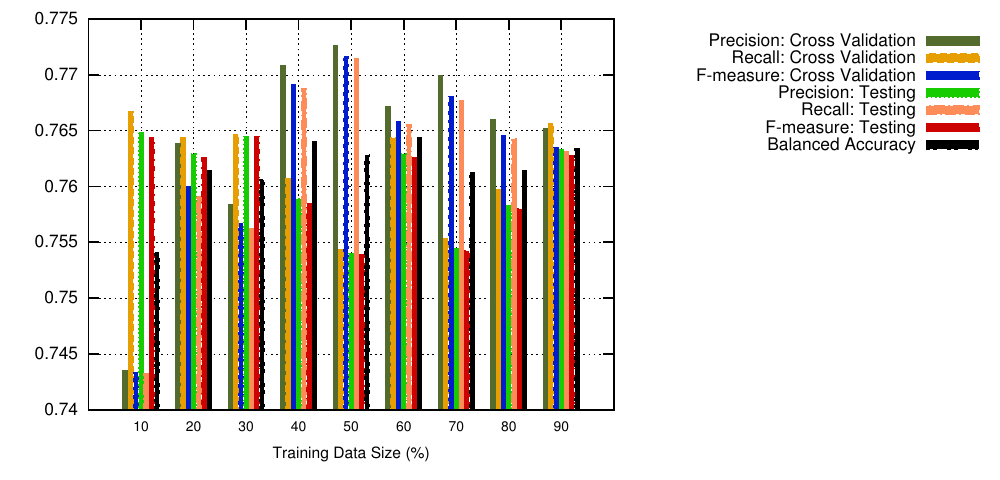}}
\caption{Performance analysis of Farthest Neighbor Histogram Metric with varying data size}
\label{fig:CatAnalysis}
\end{figure}

\begin{figure}[H]
\centerline{\includegraphics[height=7cm]{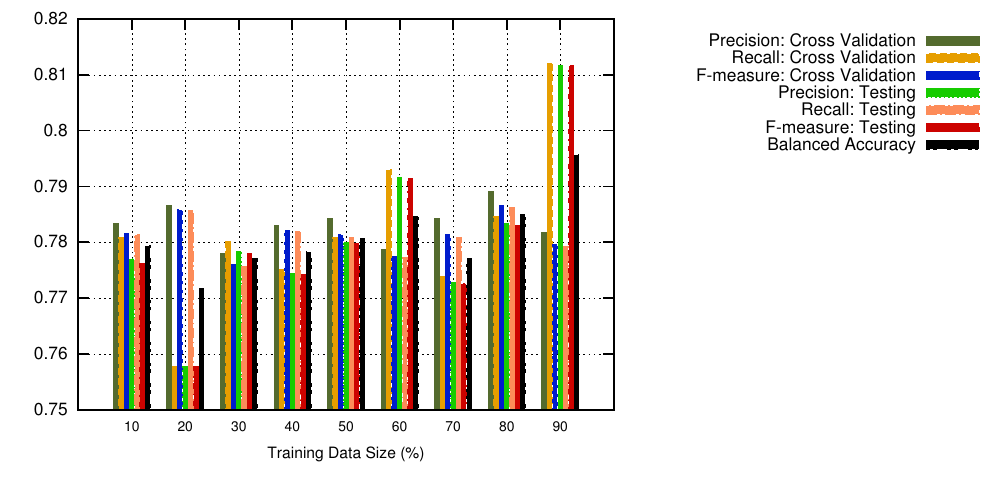}}
\caption{Performance analysis of Ranked Histogram with varying data size}
\label{fig:CatAnalysis}
\end{figure}

\begin{figure}[H]
\centerline{\includegraphics[height=7cm]{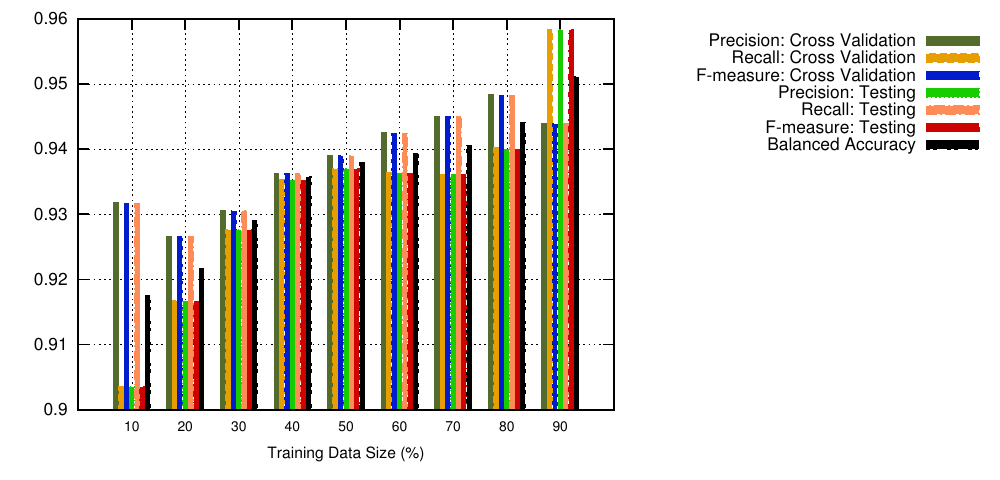}}
\caption{Performance analysis of HSV Histogram with varying data size}
\label{fig:CatAnalysis}
\end{figure}

\begin{figure}[H]
\centerline{\includegraphics[height=7cm]{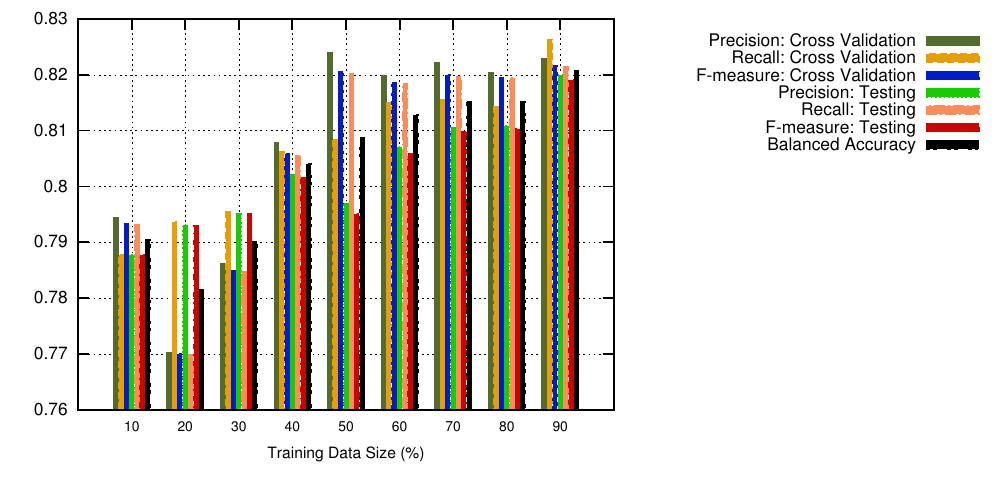}}
\caption{Performance analysis of concatenated features with varying data size}
\label{fig:CatAnalysis}
\end{figure}

The concatenated features have a balanced accuracy of 96.57\%. The HSV histogram 
and the farthest neighbor metric fared the best and worst in individual 
performance with balanced accuracies of 93.23\% and 59.86\% respectively.

\subsection{ELM Performance Analysis}
As mentioned previously, the ELM was trained on 3000 feature vectors of each class, 1320-dimensional each. The input weight matrix and the bias matrix were composed of values chosen randomly between -3 and 3. The following figure plots the F-measure for both the classes and the overall balanced accuracy of the ELM classifier.
\begin{figure}[H]
\centerline{\includegraphics[height=7cm]{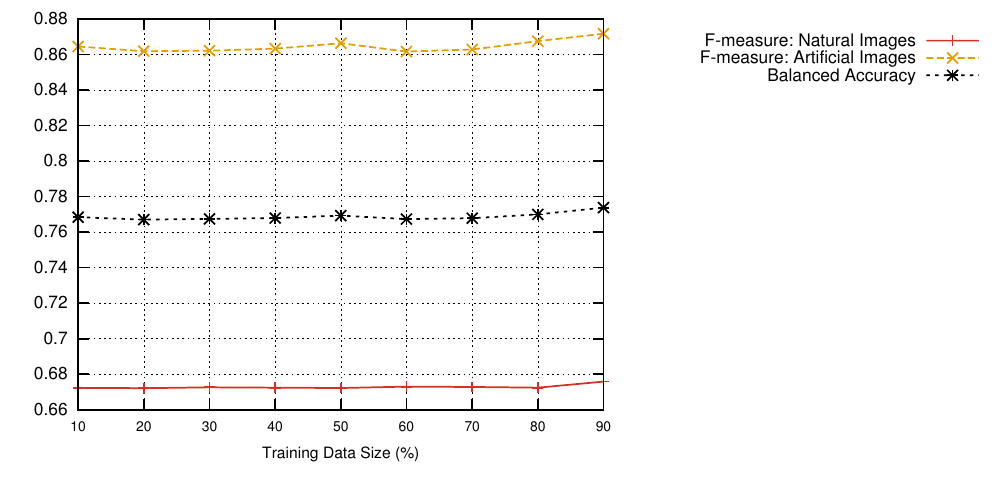}}
\caption{Performance analysis of the designed ELM classifier trained on varying dataset of varying size. The variation of F-measure of both the classes and the balanced accuracy according to the training dataset size is plotted.}
\label{fig:ELM}
\end{figure}
From the figure, it is clear that the ELM classifier is inferior to the SVM classifier for our application. Therefore, no detailed performance analysis was done for the ELM classifier and the SVM classifer was used for classification of the extracted bands.

\chapter{Conclusion}

In our project, we have successfully come up with a robust approach to detect the format of broadcast news videos. Unlike previous work done in this field, our approach assumes no prior information about the news video format. This ensures that our approach works
efficiently for unknown news formats as well.\\

A $1320$-dimensional feature vector is proposed which is used to classify the natural and the artificial bands. A text detector is used to detect the text regions. Finally, all the bands are merged together using a three-tier hierarchical reasoning to form the final format profile.\\

The entire broadcast format is represented in the form of rectangular bands containing either text, natural shots or computer generated graphics. As the entire news information is represented only through the text bands and natural shots, by removing the redundant information from computer generated graphics and converting the text images to machine encoded text, a significant reduction in the storage space can be achieved. \\ \\

\chapter{Future Work}

Our current approach accurately detects the profile bands from the video frames. This approach can be extended so as to track the moving bands in the video with an aim to detect the band animation format. This could play a very important role in detecting news shows and also in other news video analysis tasks. Further work in this could be to detect non-rectangular and overlapping bands as well.
This work, when applied to a news video in conjunction with the output of the semantic story segmentation and video shot linking, could be used as an effective way to generate the entire news summary. As of now we are dealing with video data recorded from 4 English news channels namely, CNN IBN, Times Now, NDTV 24x7 and BBC WORLD. To measure the robustness of the final proposed approach, it could also be tested on vernacular as well as other international news channels.

\section{Improving the Classification Accuracy}
The present classifier has classification accuracy of around $95\%$ (precision, recall, F-measure and balanced accuracy) 
for both the classes. However, there are a few more features that can be incorporated 
to further improve the classifier performance.

\subsection{Color Correlogram}
A color correlogram is an effective description of the global distribution of the local 
spatial correlation of colors.
As mentioned before, because of the way natural images are acquired using a camera, 
they tend to have some inherent noise, therefore making them more noisy than 
graphics. This property can be exploited using a color correlogram.\\ \\
A color correlogram of an image is a two-dimensional matrix where.
Each entry (x, y) in the matrix stores the probabilty that the given pixel
(x, y) has the same color (or within a certain threshold) as its neighbors. 
This kind of feature is not possible to obtain using a simple color
histogram since it only gives only information about the color distribution in
an image, but no information about spatial correlation of the colors.
\begin{figure}[H]
\centerline{\includegraphics[scale=0.4]{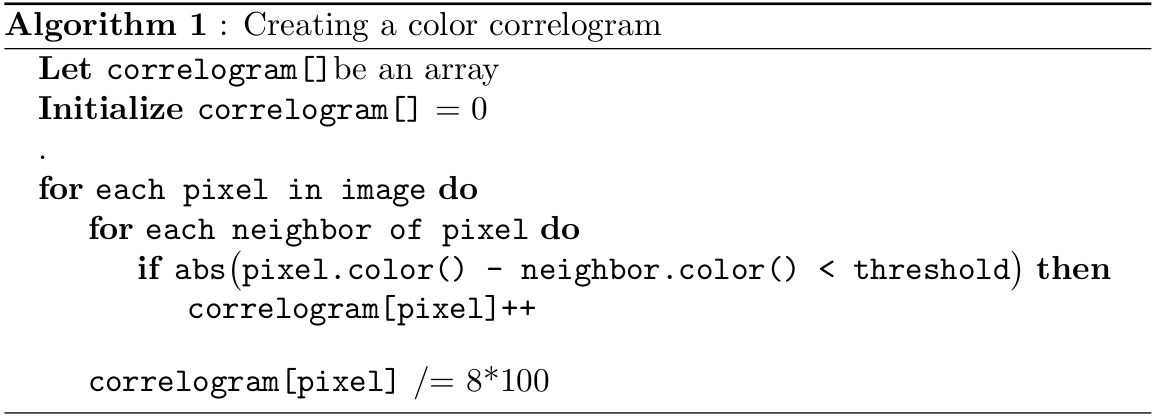}}
\label{fig:correlogram_algo}
\end{figure}

\subsection{Spatial Gray Level Dependence}
A Spatial Gray Level Dependence (SGLD henceforth) Histogram (also known as 
\textbf{Co-occurence Matrix}) is an effective representation of the image 
texture and in particular, the image sharpness. A SGLD histogram is a two-dimensional 
histogram. It denotes the joint probabilty of the occurence of two gray levels 
`i' and `j' with a defined spatial relation in terms of distance `d' and angle `$\theta$'. 
For each image, a gray level transformation is obtained, and it 
obtained by averaging the intensity values of the three R, G and B channels.
\begin{equation}
 brightness = \dfrac{red + green + blue}{3} \nonumber \\
\end{equation}
For each pixel in the image, if the pair (the pixel under consideration) and 
its neighbor belong to an area of color constancy (i.e. having the same brightness), 
an entry on the SGLD matrix diagonal is incremented by one. However, if the pair 
being considered lies on an edge (i.e. the two pixels belong to two different 
regions), the brightness levels might be similar, but not the same (in case of a 
weak edge) or they can be quite significantly different (in case of a strong/sharp 
edge). In the latter case, an entry far from the diagnonal of the SGLD matrix is 
incremented, while in the former case, an entry not so far from the diagnonal is 
incremented.

\begin{figure}[H]
\centerline{\includegraphics[scale=0.4]{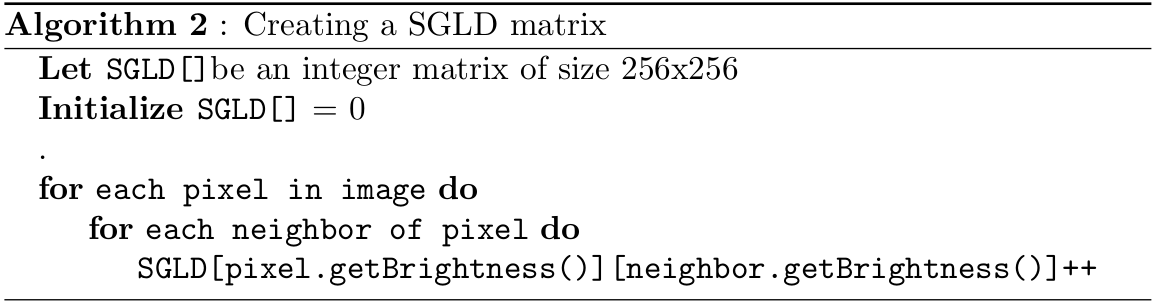}}
\label{fig:sgld_algo}
\end{figure}

\subsection{Local Binary Patterns}
Ever since their first use in 1994, Local Binary Patterns (LBPs henceforth) 
have remained a efficient feature for texture description and classification. 
Since there is a good amount of textural information that can be levaraged in 
the binary classification being dealt with here, LBPs can prove to be a 
powerful feature.\\ \\
The LBP feature vector for an image is created by comparing the pixel intensity values 
of all the neighbors of a particular pixel. A value of `0' is assigned to the 
neighbor if it has an intensity greater than that of the centre pixel, 
else a value of `1' is assigned. The assigned values are then read clockwise 
starting from the top left, and the binary number thus obtained is converted 
to decimal representation. This number is then assigned to the centre pixel. 
This process is repeated for the entire image, and has been illustrated below:

\begin{figure}[H]
\centerline{\includegraphics[scale=1.2]{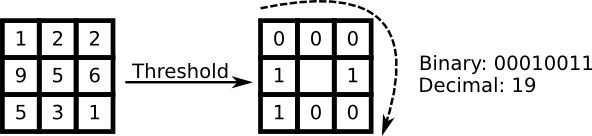}}
\caption{LBP computation considering an 8-neighborhood}
\label{fig:lbp_wiki}
\end{figure}

The algorithm for the same has been detailed below:

\begin{figure}[H]
\centerline{\includegraphics[scale=0.4]{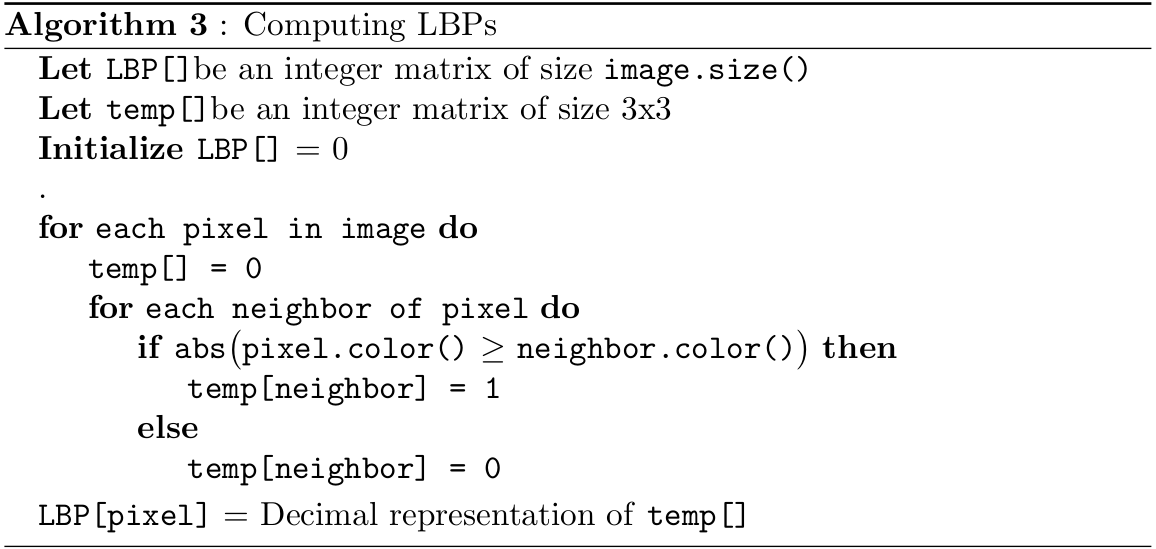}}
\label{fig:lbp_algo}
\end{figure}

\section{Better algorithm for labeling the bands}
One of the shortcomings encountered while labeling the extracted bands is that because of the bands breaking up into multiple fragments, one or more of the natural band fragments tend to be labeled as artificial. This poses a problem because since adjacent bands are assigned different labels, it is not possible to merge them. Note that the converse of this is not true, \textit{i.e.} no artificial bands are ever classified as natural.\\

\begin{figure}[!htb]
\minipage{0.32\textwidth}
  \includegraphics[width=\linewidth]{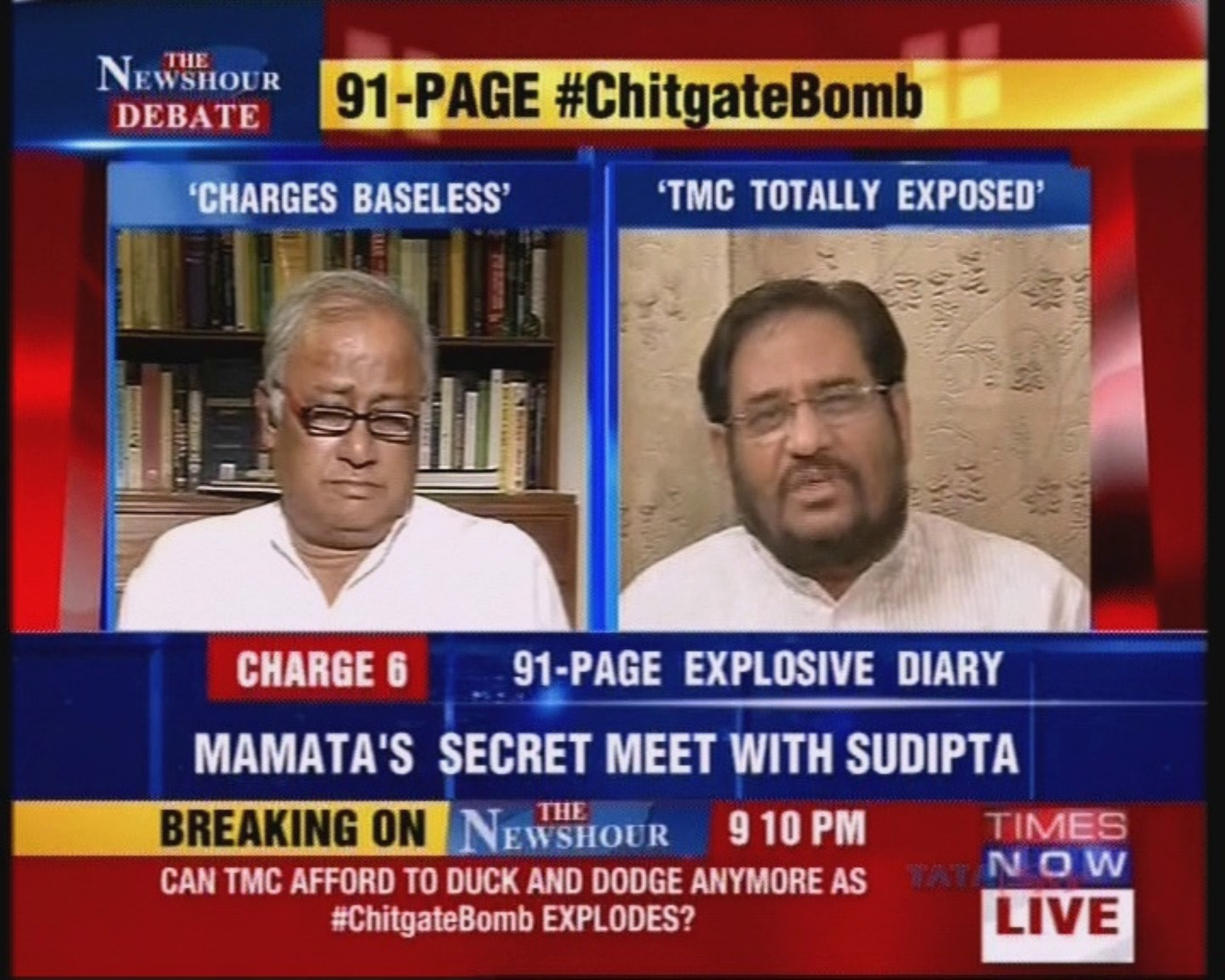}
  \subcaption{Original Image}\label{(a) katahua}
\endminipage\hfill
\minipage{0.2\textwidth}
  \includegraphics[width=\linewidth]{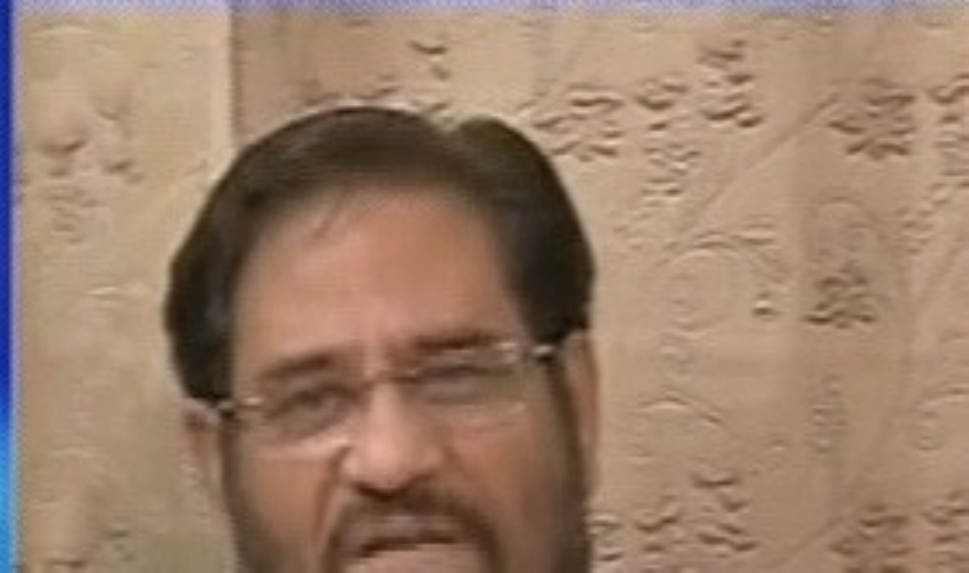}
  \subcaption{Most of the face in one image}\label{(b) hua}
\endminipage\hfill
\minipage{0.2\textwidth}%
  \includegraphics[width=\linewidth]{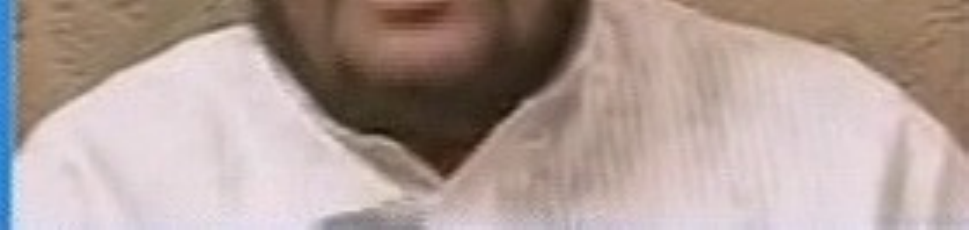}
  \subcaption{Partial face image, which gets labeled as artificial}\label{(c) kata}
\endminipage
 \caption{Single band split into multiple fragments because of the extension of Hough lines.}
  \label{fig: hoguh}
\end{figure}

A better labelling algorithm that can be tried in the future is as described below:\\
\textbf{Step 1:} Extract all the fragmented bands from a frame.\\
\textbf{Step 2:} Compute the HSV histogram of all the fragmented bands, and perform incremental clustering using the HSV histogram data as the feature vectors. This clustering of the fragmented bands based on their similarity ensures that fragments which are part of the same band get assigned to the same cluster.\\
\textbf{Step 3:} Assign labels to the bands using the classifier as usual.\\
\textbf{Step 4:} For each cluster, if any band in the cluster has been classified as natural, assign the natural label to all the bands in the cluster.\\

\chapter{Appendix}
\section{Ground Truth Marking Tool}
As the data used for training and testing of our approach is self-generated from various recorded news channels, a ground truth marking tool is developed in OpenCV, to efficiently generate the ground truth results of all the dataset images.\\

The tool loads an image and displays it on the screen. The user has to mark rectangular regions of interest with the mouse by selecting top left and bottom right corner of the rectangular bands. A text file is created for each image and after marking all the bands (both artificial and natural), the bands' positions are dumped in the respective text file. For calculating the accuracy, the result is compared with the ground truth, which is obtained by reading the text file of the respective testing image.

\section{Dataset Generating Tool}
The classifier has been trained on 3000 images of artificial and natural categories each. Since the images to be classified are the components of the news frames, the training data is also generated from the news frames. In order to achieve this, a simple tool is developed.\\

Given an image, the Hough rectangles are detected in the image as mentioned earlier. Then, a window is shown with the image and the Hough rectangles are shown on the image one by one starting from the bottom left of the image. The user provides input from the keyboard, indicating whether the rectangle currently being shown belongs to the artificial category or the natural category. We have assigned two keys, one for each class. Upon the user input, the rectangle gets dumped into the corresponding folder.\\

After the dataset has been generated, the feature vectors are generated for the images and are then used for classifier training.

\nocite{*}
\bibliographystyle{unsrt}

\end{document}